\pdfoutput=1
\documentclass[11pt]{article}

\usepackage[]{EMNLP2023}

\usepackage{times}
\usepackage{latexsym}
\usepackage{adjustbox} % for adjusting the table size
\usepackage{setspace} % for item spacing

\usepackage[T1]{fontenc}
\usepackage[utf8]{inputenc}
\usepackage{multirow}
\usepackage{microtype}
\usepackage{enumitem}

\usepackage{inconsolata}
\usepackage{subcaption}
\usepackage{amsmath}
\usepackage[normalem]{ulem}
\useunder{\uline}{\ul}{}
\newcommand\crule[3][black]{\textcolor{#1}{\rule{#2}{#3}}}
\usepackage[prependcaption,textsize=tiny]{todonotes}
\definecolor{ao}{rgb}{0.0, 0.5, 0.0}
\definecolor{orange}{rgb} {0.92156862745,0.56470588235,0.43529411764}
\definecolor{red}{rgb}{0.70588235294,0.03529411764,0.09019607843}
\definecolor{black}{rgb}{0.16862745098,0.17647058823,0.29411764705}

\usepackage{graphicx} % for including graphics

\title{DepNeCTI: Dependency-based Nested Compound Type Identification for Sanskrit}

\author{Jivnesh Sandhan\textsuperscript{$\dagger$}\thanks{* denotes the first two authors contributed equally.}, Yaswanth Narsupalli\textsuperscript{$\ddagger$}\footnotemark[1], Sreevatsa Muppirala\textsuperscript{$\ddagger$},\\ 
\textbf{Sriram Krishnan\textsuperscript{$\S$}, Pavankumar Satuluri\textsuperscript{$\|$}, Amba Kulkarni\textsuperscript{$\S$} and Pawan Goyal\textsuperscript{$\ddagger$}} \\
\textsuperscript{$\dagger$}UC Berkeley,
\textsuperscript{$\ddagger$}IIT Kharagpur,  \textsuperscript{$\S$}University of Hyderabad and
\textsuperscript{$\|$}IIT Roorkee\\
\texttt{jivneshsandhan@gmail.com, yasshu.yaswanth@gmail.com, pawang@cse.iitkgp.ac.in}
\texttt{}}

\begin{document}
\maketitle
\begin{abstract}
Multi-component compounding is a prevalent phenomenon in Sanskrit, and understanding the implicit structure of a compound’s components is crucial for deciphering its meaning. Earlier approaches in Sanskrit have focused on binary compounds and neglected the multi-component compound setting. This work introduces the novel task of nested compound type identification (NeCTI), which aims to identify nested spans of a multi-component compound and decode the implicit semantic relations between them. To the best of our knowledge, this is the first attempt in the field of lexical semantics to propose this task. 

We present 2 newly annotated datasets including an out-of-domain dataset for this task. We also benchmark these datasets by exploring the efficacy of the standard problem formulations such as nested named entity recognition, constituency parsing and seq2seq, etc. We present a novel framework named DepNeCTI: \textbf{Dep}endency-based \textbf{Ne}sted \textbf{C}ompound \textbf{T}ype \textbf{I}dentifier that surpasses the performance of the best baseline with an average absolute improvement of 13.1 points F1-score in terms of Labeled Span Score (LSS) and a 5-fold enhancement in inference efficiency. In line with the previous findings in the binary Sanskrit compound identification task, context provides benefits for the NeCTI task. 
The codebase and datasets are publicly available at: \url{https://github.com/yaswanth-iitkgp/DepNeCTI}
% \pg{If you are sharing, it should be anonymized.} \js{Sure}

\end{abstract}

\section{Introduction}
\label{introduction}
A compound is defined as a group of entities functioning as a single meaningful entity. The process of identifying the implied semantic relationship between the components of a compound in Sanskrit is known as Sanskrit Compound Type Identification (SaCTI) \cite{sandhan-etal-2022-novel} or Noun Compound Interpretation (NCI) \cite{ponkiya-etal-2021-framenet,ponkiya-etal-2020-looking}. Within the literature, the NCI problem has been approached in two ways, namely, classification \cite{dima-hinrichs-2015-automatic,fares-etal-2018-transfer,ponkiya-etal-2021-framenet} and paraphrasing \cite{lapata-keller-2004-web,ponkiya-etal-2018-treat,ponkiya-etal-2020-looking}.  
% Let's consider the compound {\sl apple juice} for illustration purposes. For instance, in the first approach, the relationship between the components is classified using a predefined set of semantic relations (e.g., "MADEOF") \cite{dima-hinrichs-2015-automatic,fares-etal-2018-transfer,ponkiya-etal-2021-framenet}. Conversely, the second approach utilizes paraphrasing to illustrate the semantic relations (e.g., "a juice made from apple") \cite{lapata-keller-2004-web,ponkiya-etal-2018-treat,ponkiya-etal-2020-looking}.

\begin{figure}[tbh]
\centering
\centerline{\includegraphics[width=2.5in]{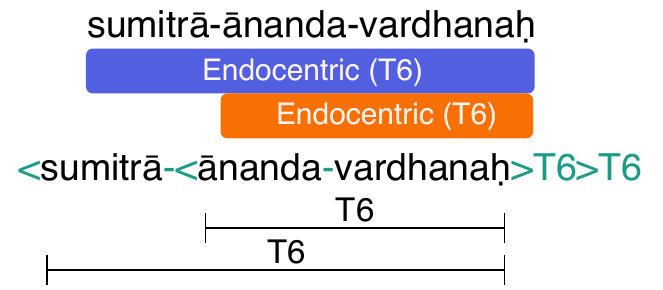}}
    \caption{Illustration of the NeCTI task for the multi-component compound \textit{sumitrā-ānanda-vardhanāḥ} (Translation: Sumitrā-delight-enhancer), highlighting nested spans and their associated semantic relations using distinct colors.}
\label{fig:task-example} 
\end{figure}

In Sanskrit literature, particularly in poetry, the use of multi-component compounds is ubiquitous \cite{anil_thesis}. According to the Digital Corpus of Sanskrit, more than 41\% of compounds contain three or more components \cite{krishna-etal-2016-compound}. However, earlier approaches focus solely on binary compounds and fail to address the complexities inherent in multi-component compounds adequately. Thus, we propose a new task, the Nested Compound Type Identification (NeCTI) Task, which focuses on identifying nested spans within a multi-component compound and interpreting their implicit semantic relationships. Figure \ref{fig:task-example} illustrates an example of the NeCTI task, highlighting nested spans and their associated semantic relations using distinct colors.

The NeCTI task presents multiple challenges:  (1) The number of potential solutions for a multi-component compound grows exponentially as the number of components increases.
(2) It often relies on contextual or world knowledge about the entities involved \cite{krishna-etal-2016-compound}.  Even if a multi-component compound shares the same components and final form, the implicit relationship between the spans can only be deciphered with the aid of available contextual information \cite{kulkarni2013,krishna-etal-2016-compound}.
\begin{figure}[tbh]
\centering
    \centerline{\includegraphics[width=3in]{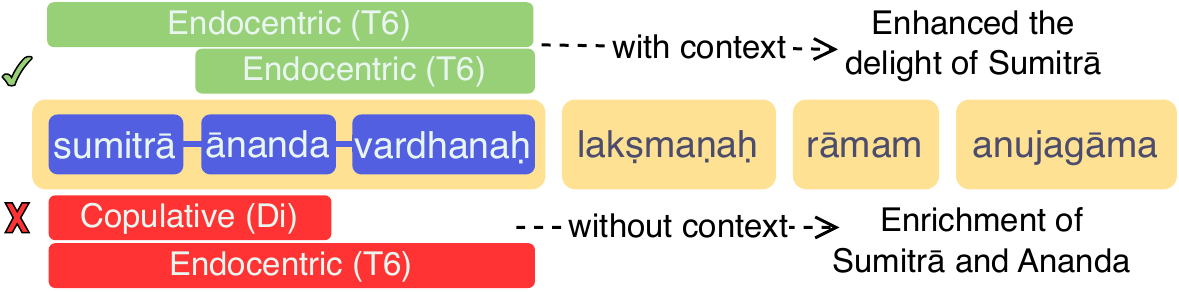}}
    \caption{Illustration of the multi-component compound \textit{sumitrā-ānanda-vardhanāḥ} (Translation: Sumitrā-delight-enhancer) with two valid parses depicted in green and red. The two parses correspond to two potential meanings. The green solution represents the correct interpretation within the provided context.}
\label{fig:example-compound} 
\end{figure}
For instance, as depicted in Figure \ref{fig:example-compound}, the multi-component compound \textit{sumitrā-ānanda-vardhanāḥ} (Translation: Sumitrā-delight-enhancer) can have two valid but distinct solutions, leading to different meanings. Resolving ambiguity to select the correct solution requires reliance on the provided context.  Consequently, downstream Natural Language Processing (NLP) applications for Sanskrit, such as question answering \cite{terdalkar-bhattacharya-2019-framework} and machine translation \cite{aralikatte-etal-2021-itihasa}, often exhibit sub-optimal performance when encountering compounds. Hence, the NeCTI task serves as a preliminary requirement for developing robust NLP technology for Sanskrit. Moreover, this dependency on contextual information eliminates the possibility of storing and conducting a lookup to identify the semantic types of nested spans.

Previous approaches \cite{kulkarni2013,krishna-etal-2016-compound,sandhan-etal-2019-revisiting} addressing SaCTI have predominantly focused on binary compounds, neglecting the consideration of multi-component compounds.
In multi-component compounds, the components exhibit semantic relationships akin to dependency relations, represented as directed labels within the dependency structure, which also facilitate the identification of the compound's headword through the labels directed towards it. Consequently, dependency formulation enables the simultaneous identification of both the structure or constituency span and the compound types.
Thus, we propose a novel framework (\S~\ref{proposed_system}) named DepNeCTI: Dependency-based Nested Compound type Identifier (\S \ref{why_dp_main}). 
% This framework (1) incorporates contextual information, and (2) offers a simplified formulation that significantly enhances efficiency with a 5-fold increase in inference speed (\S~\ref{analysis}) compared to the best-performing baseline. 
In summary, our contributions can be outlined as follows:

\begin{itemize}[leftmargin=*]
    \item We introduce a novel task called Nested Compound Type Identification (\S~\ref{proposed-task-dataset}).
    \item We present 2 newly annotated datasets and provide benchmarking by exploring the efficacy of various standard formulations for NeCTI (\S~\ref{experiments}).
    \item We propose a novel framework DepNeCTI: Dependency-based Nested Compound type Identifier (\S~\ref{proposed_system}), which reports an average 13.1 points F1-score in terms
    of LSS absolute gain and 5-fold enhancement in inference efficiency (\S~\ref{analysis}) over the best baseline.
  \item We publicly release the codebase of DepNeCTI and benchmarked baselines, along with newly annotated datasets for the NeCTI task.
\end{itemize}

\section{Problem Formulation}
\label{proposed-task-dataset} 
The objective of the NeCTI task is to detect nested spans within a multi-component compound and decipher the implicit semantic relations among them.
Our study focuses exclusively on this task and does not address the compound segmentation problem. It is assumed that the segmented components of the multi-component compound are already available. To obtain the segmentation of a compound, we rely on established resources such as the rule-based shallow parser \cite{goyal2016} or existing data-driven segmentation systems designed explicitly for Sanskrit \cite{hellwig-nehrdich-2018-sanskrit,sandhan-etal-2022-translist}.

\paragraph{Complexity of NeCTI Task:}
\begin{figure}[tbh]
\centering
    \centerline{\includegraphics[width=2.5in]{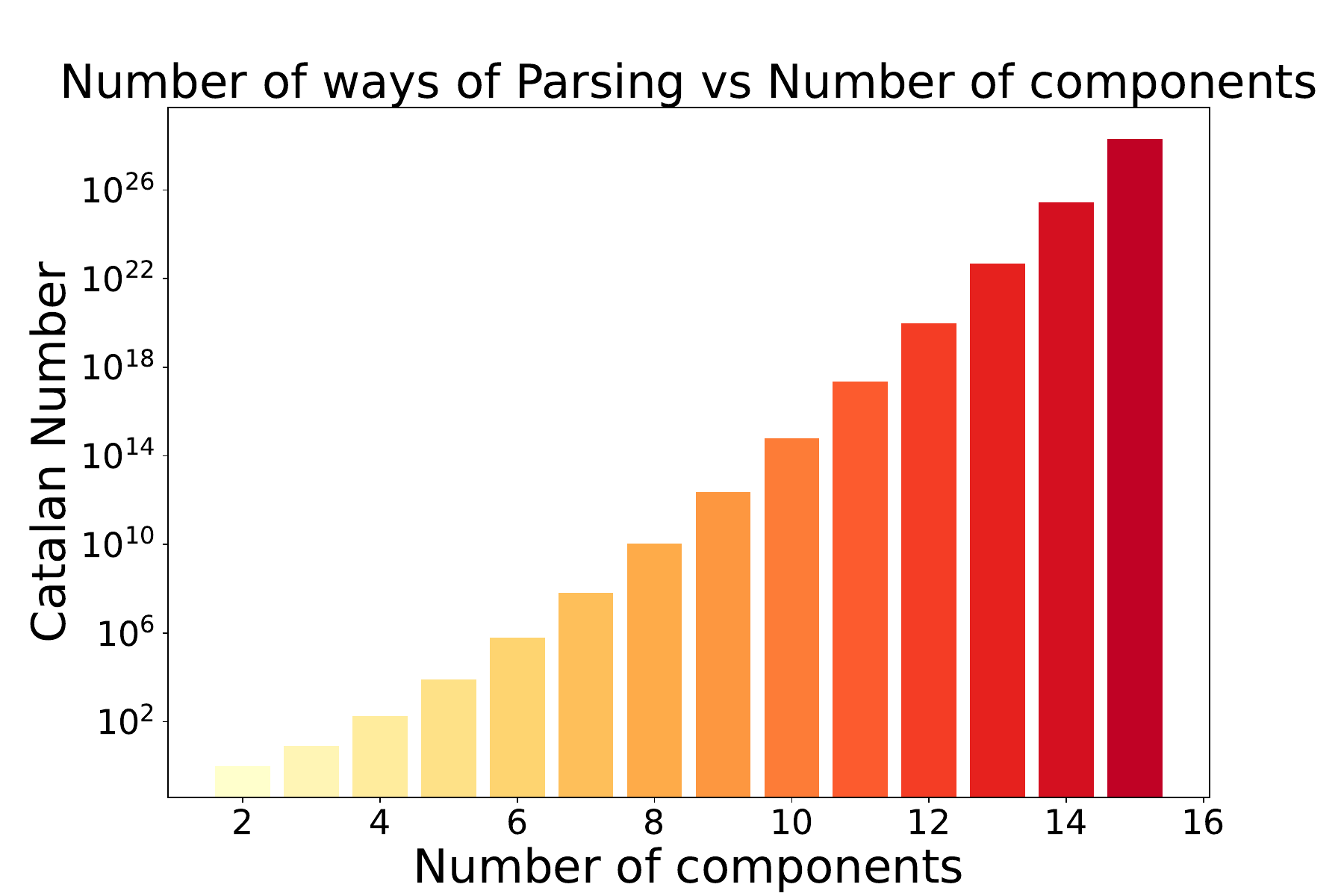}}
    \caption{Illustration of a number of ways (as per Catalan number in log scale) in which a multi-component compound can be parsed. Our dataset contains compounds that have a maximum of 16 components.}
\label{fig:exponential} 
\end{figure}

As the number of components in a multi-component compound increases, the number of possible parses grows exponentially. Our dataset encompasses multi-component compounds ranging from 2 to 16 components. Figure \ref{fig:exponential} visually depicts the exponential growth of possible parses with increasing component count. The parsing of a compound word with $n+1$ components can be likened to the problem of fully parenthesizing $n+1$ factors in all possible ways \citep{Kulkarni2011StatisticalCP}. Thus, the total number of parse-options for a multi-component compound word with $n+1$ components corresponds to the Catalan Number $C_{n}$, where $n\geq 0$ \citep{Huet2009SanskritS}. 

\[C_n = \frac{1}{{n+1}}\binom{{2n}}{{n}} = \frac{{(2n)!}}{{(n+1)!n!}}\]

where \(C_n\) represents the $n^{th}$ Catalan number, \(\binom{{2n}}{{n}}\) is the binomial coefficient, and \(!\) denotes factorial. Finally, the compatibility rules derived from P\={a}ni\d{n}\={\i}an grammar \cite{panini} and contextual information are needed to disambiguate multiple possibilities.

Formally, in a given sentence $X = \{x_1, x_2,..., x_M\}$ with $M$ tokens, let $x_p$ ($1 \leq p \leq M$) denote an $N$-component compound. It is worth noting that $X$ may contain multiple instances of multi-component compounds. A valid solution corresponds to a full paranthesization of these $N$ components; let $P_N$ encompass all valid solutions for fully parenthesizing $N$ factors, satisfying the cardinality $|P_N| = C_{N-1}$, where $C_{N-1}$ represents the Catalan number. A valid solution for an $N$-component compound consists of $N-1$ nested spans. The NeCTI system produces an output represented as a list of $N-1$ tuples for $x_p$, given by $Y_p = \{[I_1^{H}, I_1^{T}, T_1], ..., [I_{N-1}^{H}, I_{N-1}^{T}, T_{N-1}]\}$, such that $Y_p\in P_N$. $I_i^{H}$ and $I_i^{T}$ denote the head and the tail indices, respectively, of the $i^{th}$ span, and $T_i$ corresponds to the label assigned to the respective span.

% In a given sentence, let $x_p$ denote an $N$-component compound within a context consisting of $M$ tokens, represented as $X = \{x_1, x_2,..., x_M\}$, where $1 \leq p \leq M$. It is worth noting that the provided sentence may contain multiple instances of multi-component compounds.  The NeCTIS system produces an output represented as a list of tuples for the corresponding $x_p$ compound, $Y_p = \{< I_1^{H}, I_1^{T}, T_1 >, ..., < I_{N-1}^{H}, I_{N-1}^{T}, T_{N-1}>\}$, conforming to the set $P_N$. Here, $P_N$ encompasses all valid solutions for fully parenthesizing $N$ factors, satisfying the cardinality of $|P_N| = C_N$, where $C_N$ represents the Catalan number. In this representation, $I_i^{H}$ and $I_i^{T}$ denote the head index and the tail index, respectively, of the $i^{th}$ span. Additionally, $T_i$ corresponds to the semantic category assigned to the respective span. The NeCTIS formulation allows for spans to overlap, but crossing boundaries between spans are not permitted, as implied by its name.  The number of spans for a given $N$-component compound is represented by $N-1$.

\paragraph{How different is NeCTI compared to the Nested Named Entity Recognition (NNER) task?}
NNER is a component of information extraction that aims to identify and classify nested named entities within unstructured text, considering their hierarchical structure. In contrast, the NeCTI task focuses on identifying nested spans within a multi-component compound and decoding their implicit semantic relations. These tasks have several key differences: (1) NeCTI operates at the intra-word level, whereas NNER operates at the inter-word level, considering entities across a phrase. (2) In NeCTI, a multi-component compound can have multiple possible parses, requiring disambiguation through contextual cues and incorporating insights from P\={a}ni\d{n}\={\i}an grammar to address incompatibilities. Conversely, we could not find discussions related to these aspects in NNER literature. (3) NeCTI benefits from prior knowledge of the compound's location and segmented components. In contrast, the NNER task involves the additional challenge of identifying the location of entities within the text. Consequently, leveraging existing NNER frameworks for NeCTI is not straightforward due to their inability to provide explicit support for providing the location of compounds. Therefore, NeCTI presents unique characteristics and challenges that differentiate it from the NNER task, requiring specialized approaches tailored to its specific requirements.

\paragraph{How different is NeCTI compared to the Multi-Word Expressions (MWE)?}  MWEs encompass various categories such as idioms (e.g., \textit{kick the bucket}), named entities (e.g., \textit{World Health Organization}), and compounds (e.g., \textit{telephone box}), etc. Sanskrit compounds exhibit similarities to multi-word expressions, particularly multi-word compounds (nominal, noun, and verb), with respect to characteristics like collocation of components based on semantic relations between them and strict preference for ordering of the components.

    Multi-word compounds in various languages involve adjacent lexemes juxtaposed with potential semantic relations. In contrast, Sanskrit compounds are intuitively constructed based on the semantic compatibility of their components. Additionally, in Sanskrit, compounds always appear as a single word, requiring mandatory \textit{sandhi} (euphonic transformations) between their components. Conversely, certain categories of MWEs (idioms, complex function words, verb-particles, and light-verbs) have relatively fixed structures, predominantly with components separated by spaces. Furthermore, nesting within multi-word expressions is considered syntactic overlaps, while the nested structures of Sanskrit compounds result from successive combinations of components based on semantic relations, thereby clearly distinguishing Sanskrit compounds from MWEs.

\begin{figure*}[!htb]
\centering
\includegraphics[width=0.7\textwidth]{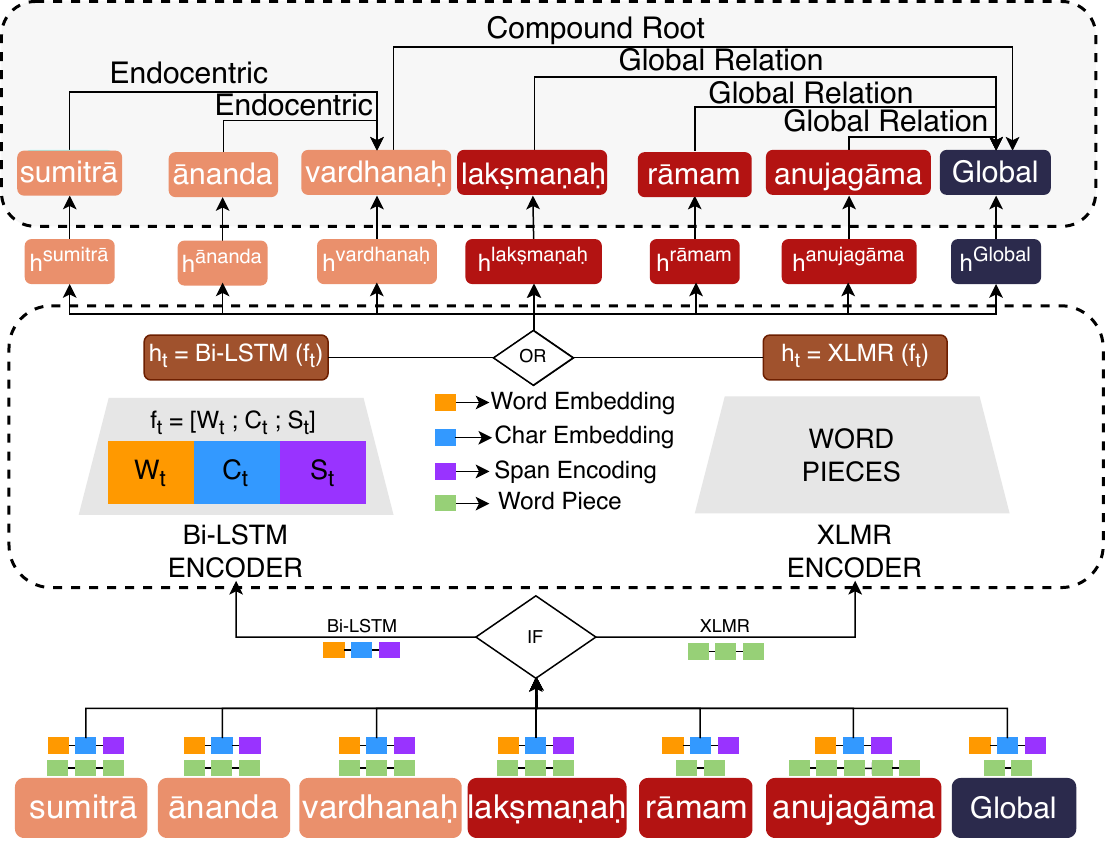}
\caption{Illustration of DepNeCTI with an example ``{\sl sumitr\={a}-\={a}nanda-vardhana\d{h} lak\d{s}ma\d{n}a\d{h} r\={a}mam anujag\={a}ma}'' (Translation: ``Lak\d{s}maṇa, the one who enhanced the delight of Sumitr\={a}, followed R\={a}ma'') where `{\sl sumitr\={a}-\={a}nanda-vardhana\d{h}}' is a multi-component compound word. We assume prior knowledge of compound segmentation and treat the individual components of multi-component compounds as separate words. That means the compound and its components are known apriori. However, the associations of the components, i.e. spans, are not known apriori. We propose two variants of DepNeCTI depending on the choice of encoder: DepNeCTI-LSTM and DepNeCTI-XLMR. To inform DepNeCTI-LSTM about compound (highlighted with \crule[orange]{0.25cm}{0.25cm} color) and non-compound (highlighted with \crule[red]{0.25cm}{0.25cm} color) tokens, we employ span encoding. The span encoding uses two randomly initialized vectors (compound or non-compound) to inform the model whether the corresponding instance is a compound or non-compound word. On the other hand, DepNeCTI-XLMR is informed about compound's location in the input string using bracketing (for example, {\sl <sumitr\={a}-\={a}nanda-vardhana\d{h}>}) and it lacks span encoding component. Next, to transform the compound-level parsing task into standard dependency parsing, we introduce (1) an additional token called ``Global'' (\crule[black]{0.25cm}{0.25cm} color) as a global head for all words in the sentence. (2) The compound head and non-compound words are connected to the Global token using the ``Compound Root'' and ``Global Relation'' relations, respectively. The hidden representations of the tokens are obtained using a Bi-LSTM or XLMR encoder. Finally, a Bi-affine \cite{dozat2017deep} dependency module is applied on top of the hidden representations.} 
\label{fig:main_model} 
\end{figure*}
\section{Why NeCTI as a Dependency Parsing Task?}
\label{why_dp_main}
The decision to formulate NeCTI as a dependency parsing task is driven by several considerations. Compounds with more than two components are typically formed through successive binary combinations, following specific semantic relations. This process creates a nested structure of binary compounds, except for a few exceptions, where the structure represents a constituency tree. However, treating this as a constituency parsing task poses challenges. The nested structure of compounds does not adhere to a syntactic structure but instead follows a semantic structure based on component relations. 
% Second, intermediate nodes within the structure are stem forms, not categories.
If we substitute the intermediate nodes with semantic relations, the same spans can be represented as dependency structures by annotating the types as directed relations. Moreover, the headwords in constituency spans are not explicitly marked but can be identified through their corresponding types. In contrast, dependency structures allow the determination of the headword based on labels directed towards it within the compound. Notably, dependency structures faithfully capture constituency information and can be mutually converted with their corresponding spans \cite{goyal-kulkarni-2014-converting}.

 Summarily, the semantic relations among compound components resemble dependency relations, which can be represented as directed labels within the dependency parse structure. This approach succinctly represents the semantic relations without introducing intermediary nodes. Lastly, it enables the simultaneous identification of the structure or constituency span alongside the identification of compound types. We encourage readers to refer to Appendix \S~\ref{Why_dp} for a more detailed illustration.

\section{DepNeCTI: The Proposed Framework}
\label{proposed_system}
Figure \ref{fig:main_model} illustrates the proposed framework with an example ``{\sl sumitr\={a}-\={a}nanda-vardhana\d{h} lak\d{s}ma\d{n}a\d{h} r\={a}mam anujag\={a}ma}'' (Translation: ``Lak\d{s}maṇa, the one who enhanced the delight of Sumitr\={a}, followed R\={a}ma'') where `{\sl sumitr\={a}-\={a}nanda-vardhana\d{h}}' is a multi-component compound word. We assume prior knowledge of compound segmentation and treat the individual components of multi-component compounds as separate words.
We propose two variants of DepNeCTI depending on the choice of encoder: DepNeCTI-LSTM and DepNeCTI-XLMR.
In DepNeCTI-LSTM, to differentiate between compound (highlighted with \crule[orange]{0.25cm}{0.25cm} color) and non-compound (highlighted with \crule[red]{0.25cm}{0.25cm} color) tokens, we employ span encoding. The span encoding uses two randomly initialized vectors (compound or non-compound) to inform the model whether the corresponding instance is a compound or non-compound word.
On the other hand, DepNeCTI-XLMR is informed about compound's location in the input string using bracketing (for example, {\sl <sumitr\={a}-\={a}nanda-vardhana\d{h}>}) and it lacks span encoding component. 
In order to convert the compound-level dependency parsing task into standard dependency parsing, we introduce two modifications. First, we introduce an additional token called ``Global'' (\crule[black]{0.25cm}{0.25cm} color) which serves as the global head for all words in the sentence. Second, we establish connections between the compound head and non-compound words to the Global token using the ``Compound Root'' and ``Global Relation'' relations, respectively. 

% Given an $N$-component compound $x_p$ in a sentence $X= [x_1, x_2, ..., x_p, ..., x_M]$ such that $p^{th}$ position $(1 \leq p \leq M)$ in the sentence is the compound word.
Formally, in a given sentence $X = \{x_1, x_2,..., x_M\}$ with $M$ tokens, let $x_p$ ($1 \leq p \leq M$) denote an $N$-component compound. Notably, X may contain multiple occurrences of multi-component compounds.
The $N$-component compound $(x_p)$ is further split into its components $(x_p = \{x_p^1,x_p^2,..., x_p^{N}\})$.  Next, we pass the overall sequence $(X = \{x_1,x_2,..., x_p^{1}, x_p^{2}, ..., x_p^{N}..., x_{M}\})$ to the encoder to obtain hidden representations. The LSTM encoder concatenates a token's word, character and span embedding to obtain its representation and the XLMR encoder uses word-pieces.   Finally, a Bi-affine \cite{dozat2017deep} dependency module is applied on top of it.

\section{Experiments}
\label{experiments}
\subsection{Datasets}
\begin{figure}[tbh]
\centering
    \centerline{\includegraphics[width=2.5in]{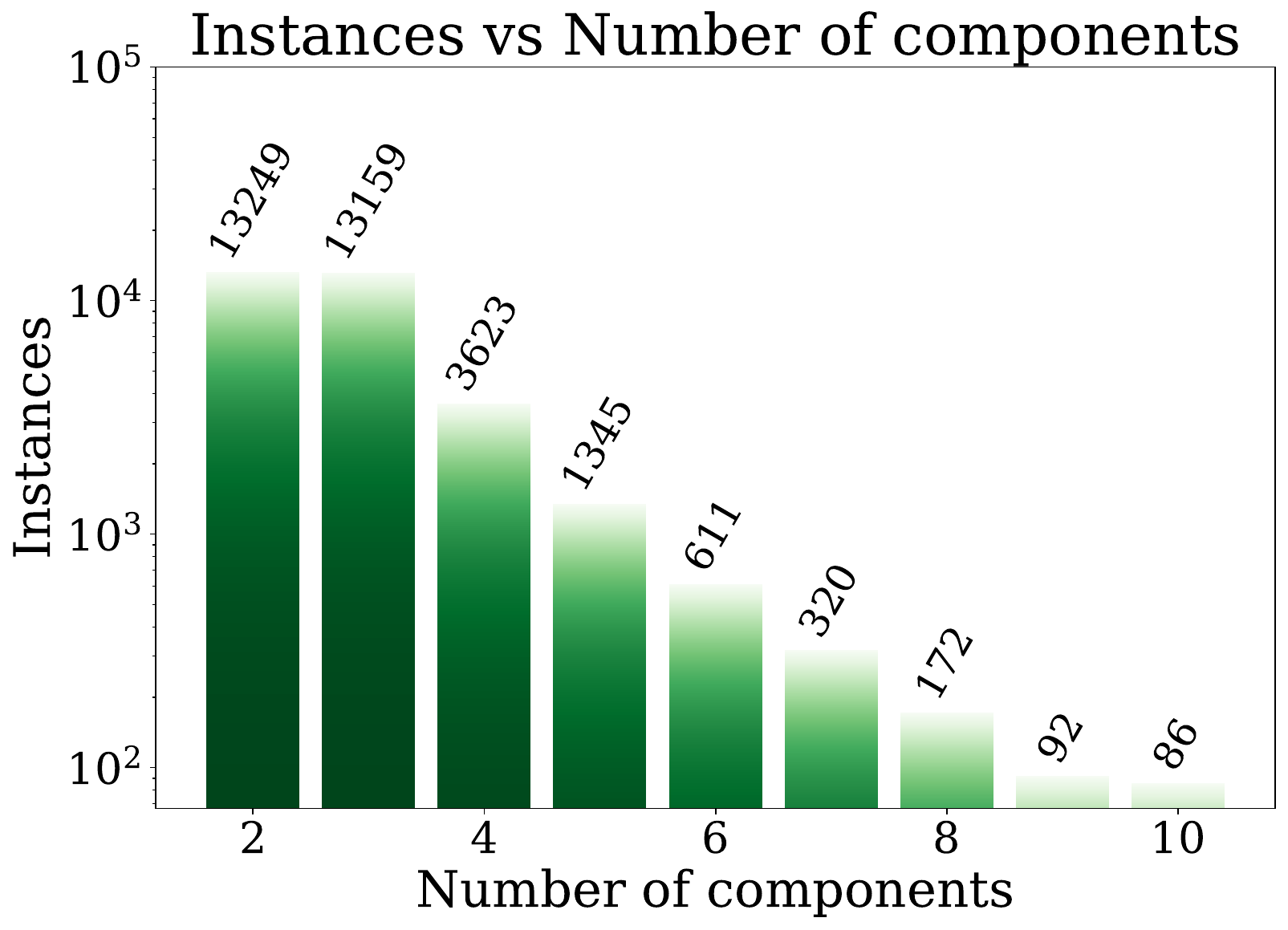}}
    \caption{Frequency of $n$-component compounds.}
\label{fig:component} 
\end{figure}

\begin{figure}[tbh]
\centering
    \centerline{\includegraphics[width=2.5in]{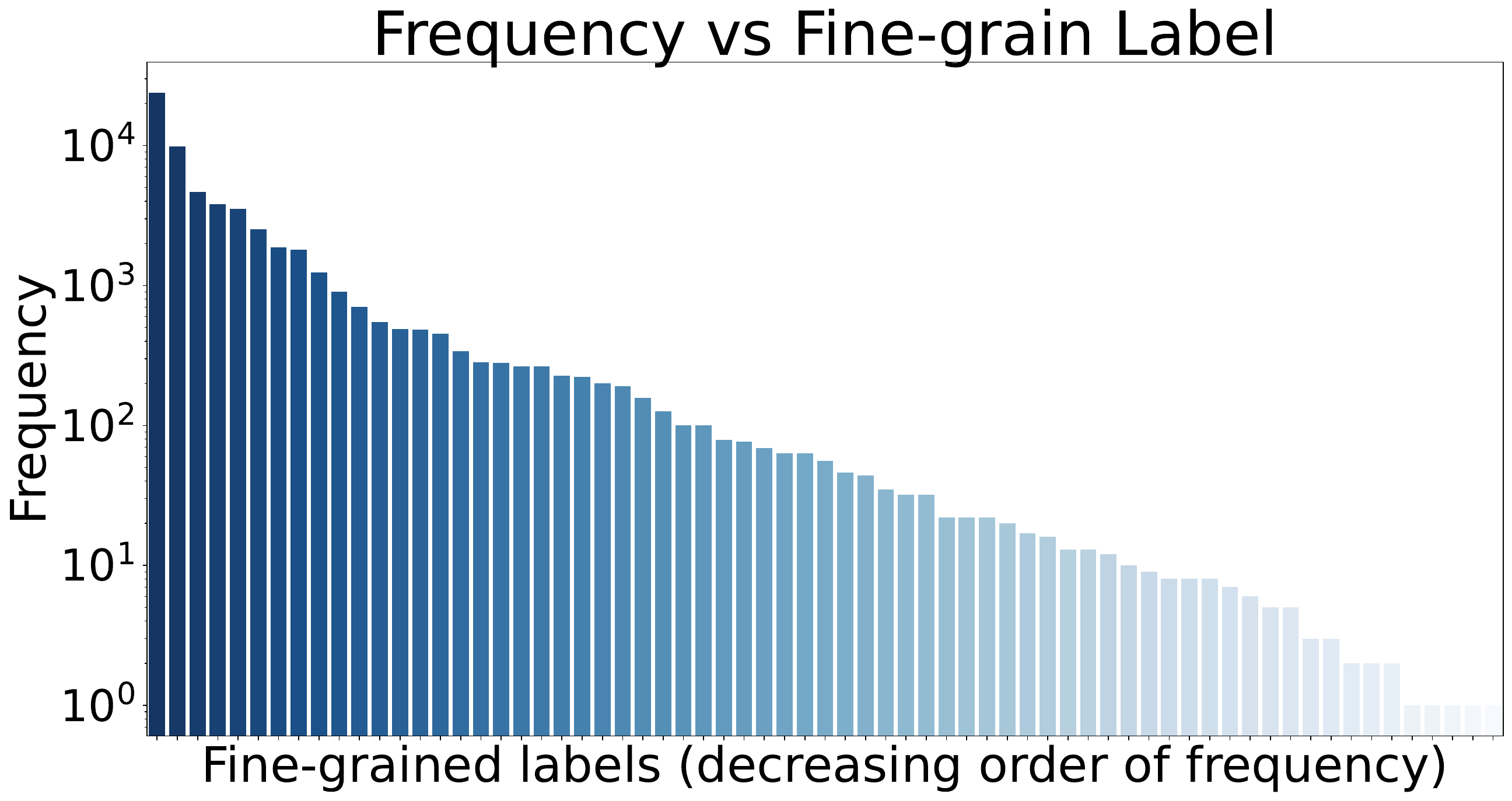}}
    \caption{Frequency in NeCTIS fine-grained.}
\label{fig:unbalanced} 
\end{figure}

Table \ref{table:dataset-statistics} presents data on the total number of multi-component compounds (Figure \ref{fig:component}) and their statistics. Our primary focus is on compounds with more than two components ($n > 2$), while also considering binary compounds if they occur in the context. These datasets comprise segmented compound components, nested spans, context, and semantic relations among the nested spans. We offer two levels of annotations for these datasets: coarse (4 broad types) and fine-grained (86 sub-types). There are 4 broad semantic types of compounds: Avyayībhava (Indeclinable), Bahuvrīhi (Exocentric), Tatpuruṣa (Endocentric) and Dvandva (Copulative). Again, each broader class is divided into multiple subclasss, leading to 86 fine-grained types.\footnote{The list of fine-grained labels and the corresponding examples can be found at: \url{https://sanskrit.uohyd.ac.in/scl/GOLD_DATA/Tagging_Guidelines/samaasa_tagging16mar12-modified.pdf}}  Figure \ref{fig:unbalanced} shows class-wise label frequency in NecTIS fine-grained.
\begin{table}[!h]
\centering
\begin{adjustbox}{width=0.45\textwidth}
\small
\begin{tabular}{|c|c|c|c|c|c|}
\hline
\textbf{Datasets}                           & \textbf{\#Nested}     & \textbf{\#Train}     & \textbf{\#Test}  & \textbf{\#Dev}  &\textbf{\#Types}  \\ \hline
NeCTIS              & 17656          & 12431        & 3493      & 2405 &4 (86)\\ \hline
NeCTIS OOD          & 1189           & $-$            & 1189         & $-$&4 (86)\\ \hline

\end{tabular}
\end{adjustbox}
\caption{Data statistics for NeCTIS and NeCTIS-OOD}
\label{table:dataset-statistics}
\end{table}

We introduce two context-sensitive datasets: NeCTIS and NeCTIS OOD. The purpose of the additional dataset (NeCTIS OOD) is to create an out-of-domain testbed. The multi-component compound instances are extracted from various books categorized into 4 types based on subject content: philosophical, paur\={a}\d{n}ic (Translation: Epic is a genre of ancient Indian literature encompassing historical stories, traditions, and legends.), literary, and \={a}yurved\={a}. The NeCTIS dataset encompasses compounds from books falling under the Philosophical, Literary, and \={a}yurved\={a} categories.  The multi-component compound instances extracted from the paur\={a}\d{n}ic category are included in the NeCTIS out-of-domain (NeCTIS OOD) dataset. Furthermore, the multi-component instances in NeCTIS belong to the prose domain, while NeCTIS-OOD pertains to the poetry domain.
Poetry commonly uses multi-component compounding extensively (more exocentric compounds) to adhere to metrical constraints and convey complex concepts. Conversely, prose uses compounds in a more direct and less condensed manner. Furthermore, poets in the realm of poetry often enjoy the freedom to form novel compounds or employ unconventional ones to conform to meter requirements, rendering these compounds infrequent in regular usage.

\begin{table*}[tbh]
\centering
\resizebox{\textwidth}{!}{%
\begin{tabular}{ccccc|ccc|ccc|ccc}
\hline
\multicolumn{1}{l}{} & \multicolumn{1}{l}{} & \multicolumn{6}{c}{{\color[HTML]{333333} \textbf{Coarse}}} & \multicolumn{6}{c}{{\color[HTML]{333333} \textbf{Fine Grain}}} \\ \cline{2-14}
\multicolumn{1}{l}{} & \multicolumn{1}{l}{} & \multicolumn{3}{c}{{\color[HTML]{333333} \textbf{w/o context}}} & \multicolumn{3}{c}{{\color[HTML]{333333} \textbf{w/ context}}} & \multicolumn{3}{c}{{\color[HTML]{333333} \textbf{w/o context}}} & \multicolumn{3}{c}{{\color[HTML]{333333} \textbf{w/ context}}} \\ \cline{1-14}
\multicolumn{1}{l|}{} & \multicolumn{1}{c|}{\textbf{Models}} & \textbf{USS} & \textbf{LSS} & \textbf{EM} & \textbf{USS} & \textbf{LSS} & \textbf{EM} & \textbf{USS} & \textbf{LSS} & \textbf{EM} & \textbf{USS} & \textbf{LSS} & \textbf{EM} \\ \hline
\multicolumn{1}{c|}{} & \multicolumn{1}{c|}{BotCP} & 72.90 & 58.78 & 32.26 & 76.22 & 63.97 & 35.10 & 74.28 & 33.50 & 18.58 & 75.72 & 41.80 & 23.05 \\
\multicolumn{1}{c|}{} & \multicolumn{1}{c|}{CP} & 76.83 & 61.71 & 39.97 & 64.73 & 46.27 & 30.14 & 77.38 & 41.86 & 27.22 & 70.56 & 32.26 & 21.11 \\
\multicolumn{1}{c|}{} & \multicolumn{1}{c|}{LexCP} & 93.39 & 84.74 & 72.88 & 93.39 & 85.16 & 74.41 & 88.70 & 45.86 & 14.72 & 87.86 & 48.87 & 19.67 \\
\multicolumn{1}{c|}{} & \multicolumn{1}{c|}{Seq2seq} & 92.54 & 84.11 & 59.90 & 91.18 & 80.45 & 52.89 & 92.67 & 65.63 & 30.65 & 92.94 & 68.19 & 34.35 \\
\multicolumn{1}{c|}{} & \multicolumn{1}{c|}{SpanCL} & 92.84 & 84.80 & 69.12 & 93.13 & 84.74 & 69.67 & 93.12 & 69.38 & 52.17 & 92.69 & 68.05 & 50.82 \\ \cline{2-14} 
\multicolumn{1}{c|}{\multirow{-6}{*}{\textbf{NeCTIS}}} & \multicolumn{1}{c|}{\textbf{DepNeCTI-LSTM}} & {\ul 95.46} & {\ul 89.06} & {\ul 76.82} & \textbf{97.42} & {\ul 89.24} & {\ul 77.00} & {\ul 95.38} & {\ul 79.49} & {\ul 57.46} & \textbf{97.49} & {\ul 79.72} & {\ul 56.83} \\
\multicolumn{1}{c|}{} & \multicolumn{1}{c|}{\textbf{DepNeCTI-XLMR}} & \textbf{96.21} & \textbf{90.83} & \textbf{79.85} & {\ul 96.16} & \textbf{90.67} & \textbf{79.45} & \textbf{96.35} & \textbf{83.36} & \textbf{63.92} & {\ul 96.34} & \textbf{83.19} & \textbf{63.30} \\
\hline
\multicolumn{1}{c|}{} & \multicolumn{1}{c|}{BotCP} & 72.81 & 48.08 & 21.63 & 73.89 & 51.18 & 22.57 & 72.30 & 20.52 & 8.64 & 73.87 & 30.10 & 13.32 \\
\multicolumn{1}{c|}{} & \multicolumn{1}{c|}{CP} & 71.43 & 52.65 & 34.57 & 68.17 & 38.63 & 25.15 & 76.12 & 29.68 & 19.77 & 69.57 & 24.10 & 15.43 \\
\multicolumn{1}{c|}{} & \multicolumn{1}{c|}{LexCP} & 89.90 & 71.44 & 50.00 &  91.57 & 72.60 & 52.73 & 83.45 & 32.85 & 7.53 & 83.38 & 34.18 & 8.31 \\
\multicolumn{1}{c|}{} & \multicolumn{1}{c|}{Seq2Seq} & 84.26 & 71.71 & 45.33 & 90.13 & 71.41 & 44.00 & 92.16 & 53.55 & 24.51 & 92.87 & 53.61 & 24.96 \\
\multicolumn{1}{c|}{} & \multicolumn{1}{c|}{SpanCL} & 91.51 & 72.69 & 56.30 & 90.24 & 71.89 & 54.46 & 89.19 & 50.21 & 32.66 & 90.88 & 50.50 & 34.00 \\ \cline{2-14} 
\multicolumn{1}{c|}{\multirow{-6}{*}{\textbf{NeCTIS OOD}}} & \multicolumn{1}{c|}{\textbf{DepNeCTI-LSTM}} & {\ul 93.32} & {\ul 78.94} & {\ul 57.40} & \textbf{95.67} & {\ul 79.26} & {\ul 57.90} & {\ul 93.88} & {\ul 67.96} & {\ul 36.60} & \textbf{95.88} & {\ul 67.26} & {\ul 36.26} \\
\multicolumn{1}{c|}{} & \multicolumn{1}{c|}{\textbf{DepNeCTI-XLMR}} & \textbf{95.56} & \textbf{84.24} & \textbf{65.50} & {\ul 95.56} & \textbf{84.45} & \textbf{65.00} & \textbf{95.56} & \textbf{74.26} & \textbf{45.70} & {\ul 95.45} & \textbf{73.54} & \textbf{44.37} \\
\hline
\end{tabular}%
}
\caption{Evaluation on the NeCTIS datasets, considering 2 levels of annotations (coarse and fine-grained) and in 2 settings (with and without context). The best-performing results are in bold, while the second-best results are underlined. The results are averaged over 3 runs. To assess the significance between the proposed system and the best baselines for each setting, a significance test in Accuracy metrics was conducted: $p < 0.01$ (as per t-test).}
\label{tab:Results}
\end{table*}

\paragraph{Dataset Annotation Process:}
% \ak{You may mention SHMT project, sponsored by DeitY, 2009-2012} 
We established a data creation process to address the unavailability of annotated context-sensitive multi-component compound data in Sanskrit. We employ a sufficient annotation budget sponsored by DeitY, 2009-2012 for the Sanskrit Hindi Machine Translation project to employ 6 institutes, each consisting of approximately 10 team members. Each team was organized in a hierarchical manner. There were 3 levels in the hierarchy: Junior linguist (Masters degree in Sanskrit), Senior linguist (Ph.D. in Sanskrit) and professional linguist (Professor in Sanskrit). The annotations from lower expertise were further checked as per the above-mentioned hierarchy. Subsequently, the annotated data underwent an exchange process with another team for correctness verification. Any ambiguities encountered during the annotation process were resolved through collective discussions conducted by the correctness-checking team. The available books were distributed among these teams, and each team was responsible for annotating their allocated books. 
The annotation guidelines\footnote{The guidelines are available at: \url{https://sanskrit.uohyd.ac.in/scl/GOLD_DATA/Tagging_Guidelines/samaasa_tagging16mar12-modified.pdf}} are essentially based on Sanskrit grammar which provides the syntactic and semantic criteria for annotation. 
% Annotators were provided comprehensive guidelines including (1) Details of coarse and fine-grained labels (2) illustrative examples of each tag (3) how to perform compound segmentation (4) how to tag nested information along with its tag (5) clarification on frequently asked questions on Sanskrit compounding.

Elaborate commentaries accompany the majority of the texts, that discuss the semantics associated with the compounds, which are typically studied by students as a part of their coursework. Given these considerations, it is very unlikely for professional linguists, often professors instructing these texts, to make mistakes. The dataset was curated around 12 years ago, primarily with the aim of producing error-free gold-standard data. Consequently, the errors made by junior annotators were not recorded or measured, aligning with our focus on achieving error-free quality. The benchmark for determining correctness was based on the P\={a}ni\d{n}\={\i}an grammar.
% Notably, all annotators had a minimum academic qualification for a Master in Arts in Sanskrit. Additionally, some annotators held Ph.D. degrees and served as professors in the field of Sanskrit.

\subsection{Baselines}
We investigate the efficacy of various standard formulations (originally proposed for nested named entity recognition for English) for the proposed task.  We adapt these systems to the NeCTI task by providing the location of the compounds to ensure a fair comparison with DepNeCTI.  Since these baselines are leveraged from the nested named entity recognition task, they do not have explicit channels to provide the location of a compound word. Therefore, we provide this information in the input string itself with the help of brackets (for example, {\sl <sumitr\={a}-\={a}nanda-vardhana\d{h}>}):
\begin{itemize}[leftmargin=*]
\item \textbf{Constituency Parsing (CP):} Following \newcite{fu2020nested}, we formulate NeCTI as constituency parsing with partially-observed trees, with all labeled compound spans as observed nodes in a constituency tree and non-compound spans as latent nodes. We leverage TreeCRF to model the observed and the latent nodes jointly.
\item \textbf{Bottom-up Constituency Parsing (BotCP):} Following \newcite{yang2021bottomup}, we formulate NeCTI as a bottom-up constituency parsing, where a pointer network is leveraged for post-order traversal within a constituency tree to enhance parsing efficiency, enabling linear order parsing.
\item \textbf{Span Classifier (SpanCL):} Following \newcite{yuan2022fusing}, we formulate NeCTI as a span classification problem, where triaffine mechanism is leveraged to learn a better span representation by integrating factors such as inside tokens, boundaries, labels, and related spans.
\item \textbf{Lexicalized Constituency Parsing (LexCP):} Following \newcite{lou-etal-2022-nested}, we formulate NeCTI as lexicalized constituency parsing, which embeds a constituency and a dependency trees together. This formulation leverages the constituents' heads into the architecture which is crucial for the NeCTI task.  
\item \textbf{Seq2Seq:} Following \newcite{yan2021unified}, we formulate NeCTI as  an entity span sequence generation task using the pretrained seq2seq framework. This generative framework can also identify discontinuous spans; however, NeCTI does not have such instances.

\item \textbf{DepNeCTI:} We propose two variants of our system (\S \ref{proposed_system}) depending on the choice of encoders: DepNeCTI-LSTM and DepNeCTI-XLMR \cite{nguyen-etal-2021-trankit}. The compound's location is provided in DepNeCTI-LSTM using span encoding; however, DepNeCTI-XLMR lacks span encoding and leverages this information from the input similar to other baselines.

% \item \textbf{DepNeCTI-XLMR:} We replace the default word embeddings in our system (\S \ref{proposed_system}) with a multi-lingual XLMR \cite{nguyen-etal-2021-trankit} and provide the compound's location in the input similar to other baselines. It lacks span encoding component.
\end{itemize}

\paragraph{Evaluation Metrics:}
We evaluate the performance using the Labeled/Unlabeled Span Score (LSS/USS) in terms of micro-averaged F1-score. We define LSS as a micro-averaged F1-score applied on tuples of predicted spans including their labels. We exclude labels of the spans while calculating USS. Additionally, we report the exact match (EM) which indicates the percentage of the compounds for which the predictions of all spans and their semantic relations are correctly identified. Refer to Appendix \ref{details} for hyper-parameters and details of the computing infrastructure used.
% To ensure consistency, the same evaluation script is used to compute these metrics across all baselines.

\subsection{Results}
\label{results}
Table~\ref{tab:Results} presents the performance of the top-performing configurations among all baselines on the NeCTIS benchmark datasets' test set. The evaluation includes 2 levels of annotations (coarse and fine-grained) and in 2 settings (with and without context). While all baseline systems demonstrate competitiveness, no single baseline consistently outperforms the others across all settings. Consequently, we underline the best-performing numbers within each specific setting.

Our proposed system DepNeCTI surpasses all competing systems across all evaluation metrics, demonstrating an absolute average gain of $13.1$ points (LSS) and $11.3$ points (EM)  compared to the best-performing baseline in each setting. 
% According to the formulation, our system accurately predicts the number of spans equal to the gold number of spans ($N-1$ spans for a $N$-component compound)), resulting in identical P, R, and F metrics. 
Notably, our proposed system exhibits substantial performance superiority over the best baseline in fine-grained settings, particularly in low-resourced scenarios. This validates the effectiveness of our proposed system in low-resourced settings with fine-grained labels. The significant performance gap between our proposed system and the best baselines highlights the efficacy of employing a simple yet effective architecture inspired by the dependency parsing paradigm. These results establish new state-of-the-art benchmarks by integrating the contextual component into our novel framework.
While most baselines (except BotCP) do not benefit from contextual information, DepNeCTI-LSTM demonstrates slight improvements and DepNeCTI-XLMR shows on par improvements when leveraging contextual information. 
Furthermore, as the number of components grows, the number of potential solutions increases exponentially, leading to poor performance by the systems in such scenarios. Due to this exponential possibility, contextual information provides limited improvements compared to binary compound identification settings \cite{sandhan-etal-2022-novel}. In other words, unlike the context-free setting, the introduction of context information does not warrant an expectation for the system to precisely generate the correct solution from the exponential candidate space. Figure \ref{fig:exponential} provides a visual representation that elucidates the concept of this exponential candidate space.
A similar performance trend is observed for the NeCTIS-OOD dataset.

\section{Analysis}
\label{analysis}
Here, we examine the proposed system, focusing on a comprehensive analysis and its applicability. For this purpose, we utilize the NeCTIS coarse dataset under the w/ context configuration. We report LSS in terms of macro-average F1-scores.

\paragraph{(1) Ablation analysis:} In this study, we analyze the impact of different system components on the overall enhancements of DepNeCTI-LSTM. Ablations, documented in Table~\ref{table:ablation}, present the evaluation metrics when a specific component is deactivated within DepNeCTI-LSTM. For instance, the absence of the span encoding component is denoted as ``- Span Encoding''. The results indicate that removing any component leads to a decline in performance. Notably, Table~\ref{table:ablation} highlights the significance of the ``Span Encoding'' component in driving the improvements.
\begin{table}[h]
\centering
\begin{adjustbox}{width=0.45\textwidth}
\small
\begin{tabular}{|c|c|c|c|c|}
\hline
\textbf{System}                            & \textbf{P}     & \textbf{R}  & \textbf{F1}  & \textbf{EM}      \\ \hline
DepNeCTI-LSTM               & 89.24 & 89.24 & 89.24 & 77.00  \\\hline
- FastText (FT)         & 88.84 & 88.84 & 88.84 & 76.50  \\\hline
- Span Encoding  (SE)      & 86.18 & 85.53 & 86.85 & 70.54 \\\hline
- FT - SE & 84.25 & 84.23  & 84.24 & 67.86\\\hline

\end{tabular}
\end{adjustbox}
\caption{Each ablation involved the removal of a singular component from DepNeCTI-LSTM. The ablation denoted as ``-Span Encoding'' entailed eliminating the span encoding component from the proposed system.
} 
\label{table:ablation}
\end{table}

\begin{figure}[h]
\centering
    \centerline{\includegraphics[width=2.5in]{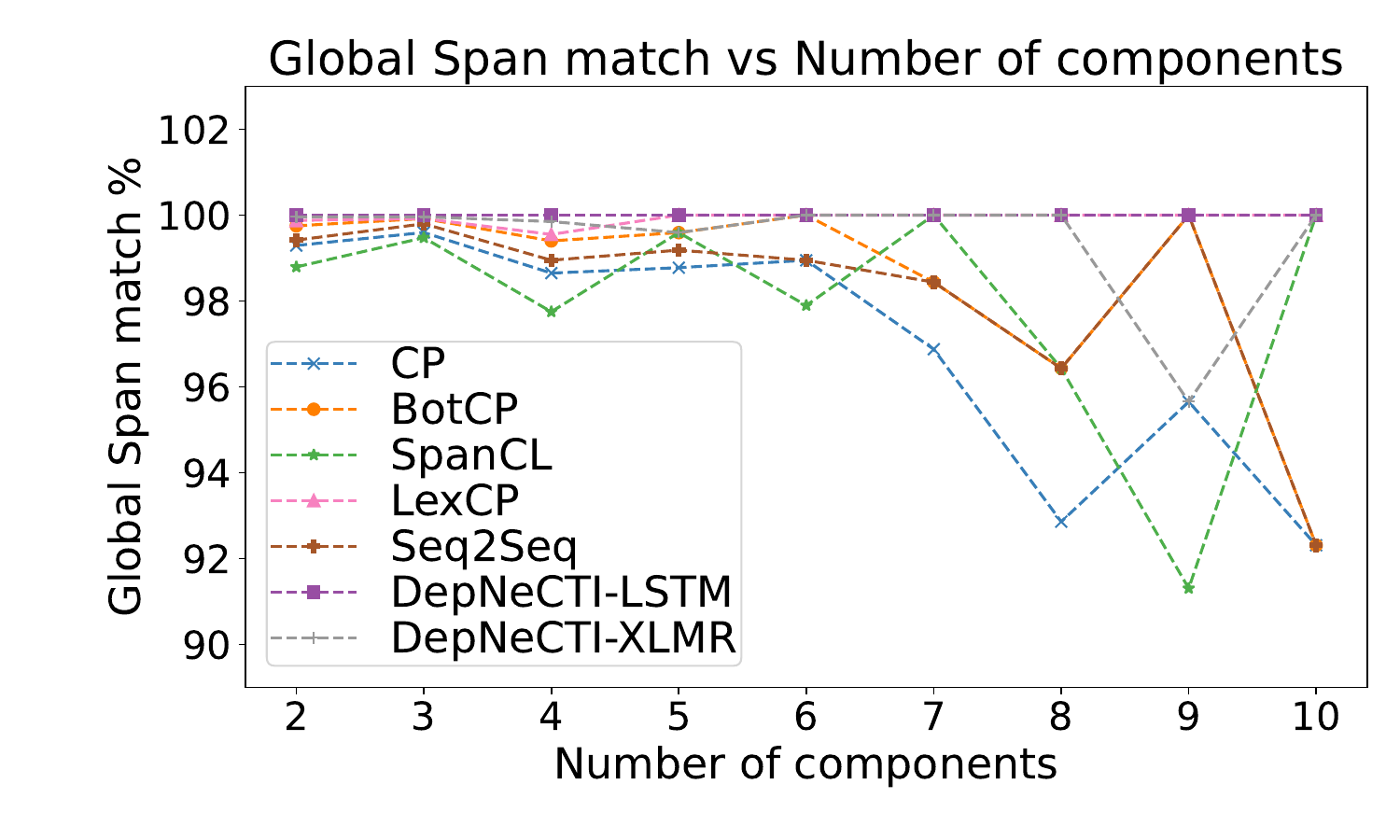}}
    \caption{Performance of the systems on identifying the global span of a compound.}
\label{fig:global_span} 
\end{figure}

\paragraph{(2) How robust is the system when the number of components of a compound increases?}
We analyze the relationship between the F1-score and the number of components in compounds. For compounds with a small number of components, all systems demonstrate high performance, but our proposed systems consistently outperforms other baselines. However, as the number of components increases, the number of examples in each category decreases. Additionally, the number of potential solutions grows exponentially, following the Catalan number. Consequently, all systems experience a decline in performance.
\begin{figure}[tbh]
\centering
    \centerline{\includegraphics[width=2.5in]{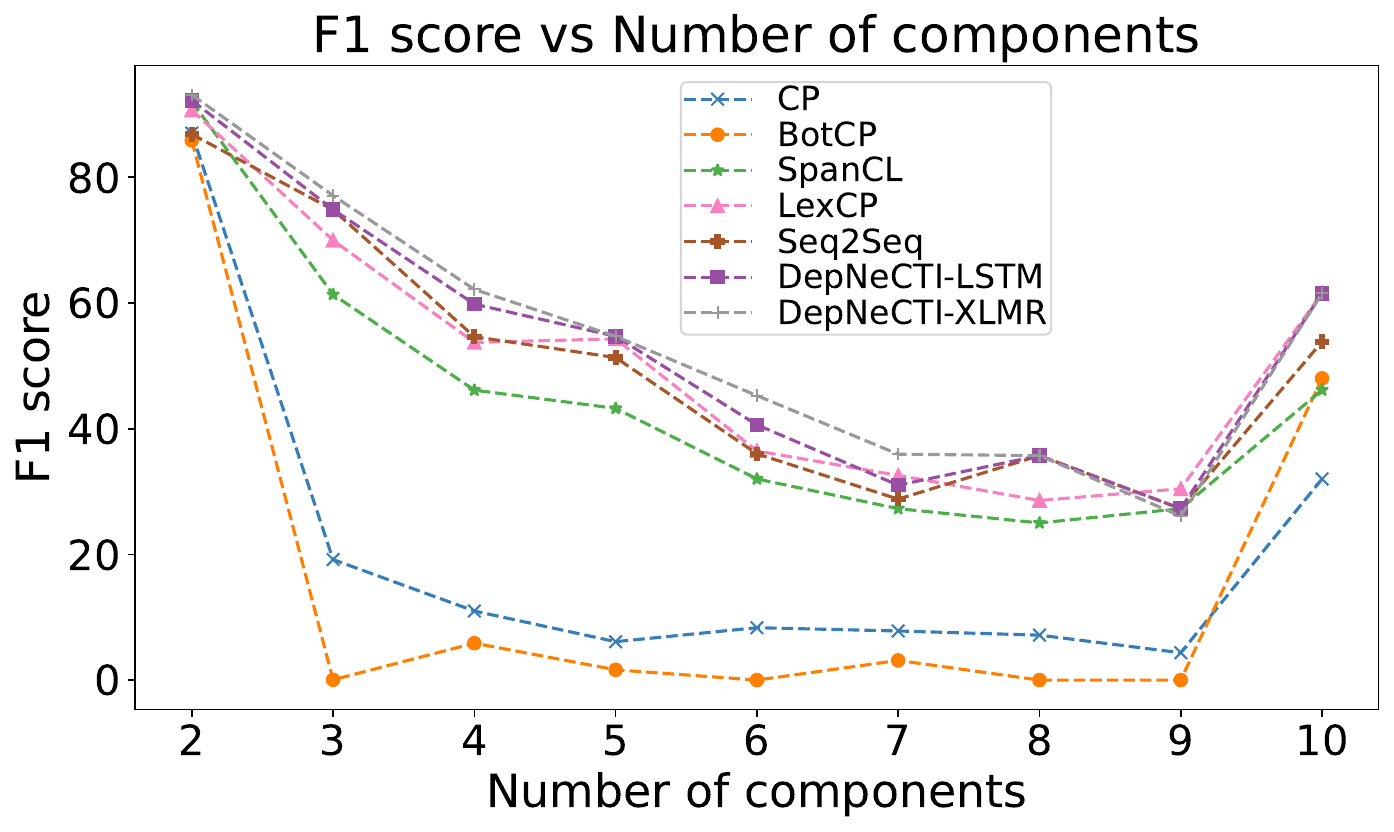}}
    \caption{F1-Score against the number of components. The compounds with components $N>10$ are excluded.}
\label{fig:f1_bucket} 
\end{figure}
\paragraph{(3) Error analysis:} 
% This section focuses on analyzing the predictions made by all systems in order to identify potential sources of errors.  Since the baselines are leveraged from the nested named entity recognition task, they do not have explicit channels to provide the location of a compound word. Thus, we have provided this information in the input string itself with the help of brackets. 
We investigate whether all the systems are able to identify the location of a multi-component compound correctly.
The motivation behind this experiment is to evaluate the capability of the baselines and the proposed architecture to leverage the information about the compound's location effectively.
We define the span of text that corresponds to a compound as a global span which we know apriori. 
Figure \ref{fig:global_span} illustrates the effectiveness of each system in correctly identifying the global span of multi-component compounds. 
 % The span of text that corresponds to a compound is already available in the input. This is defined as a global span which we know apriori. The task is to identify the correct subspans from the global span i.e. compound.
 % We want to emphasize that Figure \ref{fig:global_span} displays the performance of the global span, which we already have prior knowledge of. 
 In DepNeCTI-LSTM, our span encoding effectively captures this information, resulting in a perfect 100\% score. However, even after providing the baselines with this information, they fail to use it due to limitations in their architectures. Interestingly, DepNeCTI-XLMR does not contain a span encoding component and leverages the compound's location information from the input as provided for the baselines. Still, DepNeCTI-XLMR reports the best performance due to its powerful word representation ability.
% Notably, DepNeCTI-LSTM consistently demonstrates accurate identification of global spans as even the number of components increases, thanks to effective modeling using span encoding. However, the performance of the other systems slightly deteriorates.
It is worth noting that the NeCTIS dataset exhibits an inherent bias towards left branching, as indicated by the nested tree structure. Consequently, all systems display a bias towards left branching as well.
Therefore, due to the dominance of left-branching instances and increased variance due to less number of instances, a spike is observed in the results.

\begin{figure}[tbh]
\centering
    \centerline{\includegraphics[width=2.5in]{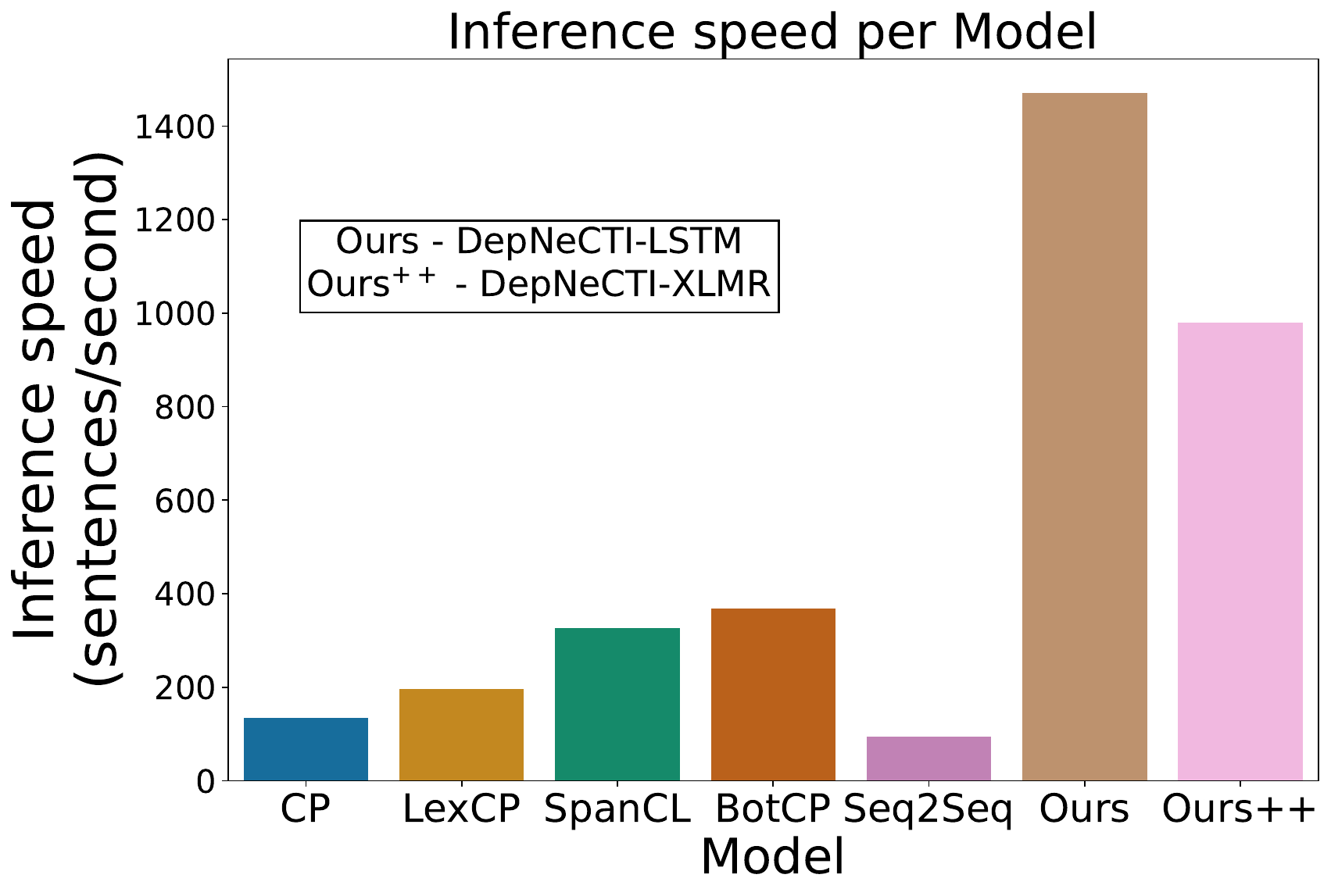}}
    \caption{Inference speed of the competing systems.}
\label{fig:inference} 
\end{figure}

\paragraph{(4) Efficiency of our proposed system:}
Figure \ref{fig:inference} present the computational efficiency of our system measured in terms of the number of sentences processed per second. We compare the inference speed of different baselines on the NeCTI task. Our systems, DepNeCTI-LSTM and DepNeCTI-XLMR are leveraging a simple architecture and utilizing dependency parsing as an appropriate problem formulation, exhibits a 5-fold/3-fold improvement over the most efficient baseline, BotCP.
\section{Related Works}
\label{related-works}
\paragraph{Lexical semantics} is a dedicated field focused on word meaning. 
% Understanding word meanings begins with syntactic analysis, where morphological examination and lemmatization offer valuable insights into the semantic roles of words within sentences \cite{kulkarni-etal-2020-dependency}. It is crucial to identify semantic relationships between words based on their senses to address ambiguities arising from word senses. 
Various tasks such as word-sense disambiguation \cite{bevilacqua-navigli-2020-breaking,barba-etal-2021-consec,10.1007/978-981-19-3148-2_40,inbook-archana}, relationship extraction \cite{tian-etal-2021-dependency,nadgeri-etal-2021-kgpool,hu-etal-2022-improving,wang2022more}, and semantic role labeling \cite{he-etal-2018-jointly,kulkarni-sharma-2019-paninian,zhang-etal-2022-label-definitions} play essential roles in determining word meaning.
Moreover, when dealing with complex word structures such as compounds, named entities, and multi-word expressions, relying solely on basic word senses and relationships is inadequate.
While efforts have been made in Noun Compound Identification \cite{ziering-van-der-plas-2015-distance,dima-hinrichs-2015-automatic,fares-etal-2018-transfer,shwartz-waterson-2018-olive,ponkiya-etal-2018-treat,ponkiya-etal-2020-looking,ponkiya-etal-2021-framenet}, multi-word expression (MWE) \cite{constant-etal-2017-survey,gharbieh-etal-2017-deep,gooding-etal-2020-incorporating,premasiri2022berts} and named entity recognition \cite{fu2020nested,yang2021bottomup,lou-etal-2022-nested,yuan2022fusing}, the nested compounds remains unexplored.

% To fill this research gap, this work introduces the NeCTI task, datasets, and framework to tackle the phenomenon of nested compounding in Sanskrit.

\paragraph{Sanskrit Compound Type Identification} has attracted significant attention over the past decade. Decoding the meaning of a Sanskrit compound requires determining its constituents \cite{Huet2009SanskritS,mittal-2010-automatic,hellwig-nehrdich-2018-sanskrit}, understanding how these constituents are grouped \cite{Kulkarni2011StatisticalCP}, identifying the semantic relationship between them \cite{anil_thesis}, and ultimately generating a paraphrase of the compound \cite{Kumar2009SanskritCP}. Previous studies proposed  rule-based approaches \cite{pavan2013,kulkarni2013}, a data-driven approach \cite{sandhan-etal-2019-revisiting} and a hybrid approach \cite{krishna-etal-2016-compound} for SaCTI.  \newcite{sandhan-etal-2022-novel} proposed a context-sensitive architecture for binary compounds.

Earlier works in Sanskrit solely focused on binary compounds, neglecting the identification of multi-component compound types; however, our proposed framework fills this research gap.

\section{Conclusion}
\label{conclusion}
In this work, we focused on multi-component compounding in Sanskrit, which helps to decode the implicit structure of a compound to decipher its meaning. While previous approaches have primarily focused on binary compounds, we introduced a novel task called nested compound type identification (NeCTI).
This task aims to identify nested spans within multi-component compounds and decode the implicit semantic relations between them, filling a gap in the field of lexical semantics.
To facilitate research in this area, we created 2 newly annotated datasets, designed explicitly for the NeCTI task. These datasets were utilized to benchmark various problem formulations. Our novel framework DepNeCTI outperformed the best baseline system by achieving a stupendous absolute gain of 13.1 points F1-score in terms of LSS. Similar to the previous findings on binary Sanskrit compound identification, we discovered that our proposed system exhibits substantial performance superiority over the best baseline in low-resourced scenarios.

\section*{Limitations}
We could not extend our framework to other languages exhibiting multi-component compounding phenomena due to the lack of availability of annotated datasets. It would be interesting to measure the effectiveness of rules from P\={a}ni\d{n}\={\i}an grammar to discard incompatible semantic relations \cite{inbook-amba,kulkarni-acm}.

\section*{Ethics Statement}
This work introduces a new task, along with annotated datasets and a framework, to address the nested compounding phenomenon in Sanskrit. The proposed resources aim to enhance the understanding of multi-component compounds and contribute to the improvement of machine translation systems.
Regarding potential effects, we anticipate no harm to any community resulting from the use of our datasets and framework. However, we advise users to exercise caution, as our system is not flawless and may generate mispredictions.
To ensure transparency and future research, we have publicly released all our annotated NeCTIS datasets and source codes. We confirm that our data collection adheres to the terms of use of the sources and respects the intellectual property and privacy rights of the original authors.
Our annotation team consisted of qualified individuals, including Master's and Ph.D. degree holders, some of whom are Sanskrit professors. Annotators were compensated appropriately and provided with detailed instructions to ensure consistency in the annotation process.
 We remain committed to addressing ethical implications as we refine our systems and welcome feedback from the community to enhance our ethical practices.

\section*{Acknowledgements}
We thank all the annotators from different institutes for helping us with NeCTIS data annotation. We would like to thank Zheng Yuan, Tsinghua University (Alibaba Group), for helping us with the hyperparameter section and adapting his system for the NeCTI task.  We appreciate and thank all the anonymous reviewers for their constructive feedback towards improving this work. The work was supported in part by the National Language Translation Mission (NLTM): Bhashini project by Government of India.

\bibliography{anthology,custom}

\begin{thebibliography}{50}
\expandafter\ifx\csname natexlab\endcsname\relax\def\natexlab#1{#1}\fi

\bibitem[{Aralikatte et~al.(2021)Aralikatte, de~Lhoneux, Kunchukuttan, and
  S{\o}gaard}]{aralikatte-etal-2021-itihasa}
Rahul Aralikatte, Miryam de~Lhoneux, Anoop Kunchukuttan, and Anders S{\o}gaard.
  2021.
\newblock \href {https://doi.org/10.18653/v1/2021.wat-1.22} {Itihasa: A
  large-scale corpus for {S}anskrit to {E}nglish translation}.
\newblock In \emph{Proceedings of the 8th Workshop on Asian Translation
  (WAT2021)}, pages 191--197, Online. Association for Computational
  Linguistics.

\bibitem[{Barba et~al.(2021)Barba, Procopio, and
  Navigli}]{barba-etal-2021-consec}
Edoardo Barba, Luigi Procopio, and Roberto Navigli. 2021.
\newblock \href {https://doi.org/10.18653/v1/2021.emnlp-main.112}
  {{C}on{S}e{C}: Word sense disambiguation as continuous sense comprehension}.
\newblock In \emph{Proceedings of the 2021 Conference on Empirical Methods in
  Natural Language Processing}, pages 1492--1503, Online and Punta Cana,
  Dominican Republic. Association for Computational Linguistics.

\bibitem[{Bevilacqua and Navigli(2020)}]{bevilacqua-navigli-2020-breaking}
Michele Bevilacqua and Roberto Navigli. 2020.
\newblock \href {https://doi.org/10.18653/v1/2020.acl-main.255} {Breaking
  through the 80{\%} glass ceiling: {R}aising the state of the art in word
  sense disambiguation by incorporating knowledge graph information}.
\newblock In \emph{Proceedings of the 58th Annual Meeting of the Association
  for Computational Linguistics}, pages 2854--2864, Online. Association for
  Computational Linguistics.

\bibitem[{Constant et~al.(2017)Constant, Eryi{\v{g}}it, Monti, van~der Plas,
  Ramisch, Rosner, and Todirascu}]{constant-etal-2017-survey}
Mathieu Constant, G{\"u}l{\c{s}}en Eryi{\v{g}}it, Johanna Monti, Lonneke
  van~der Plas, Carlos Ramisch, Michael Rosner, and Amalia Todirascu. 2017.
\newblock \href {https://doi.org/10.1162/COLI_a_00302} {{S}urvey: Multiword
  expression processing: A {S}urvey}.
\newblock \emph{Computational Linguistics}, 43(4):837--892.

\bibitem[{Dima and Hinrichs(2015)}]{dima-hinrichs-2015-automatic}
Corina Dima and Erhard Hinrichs. 2015.
\newblock \href {https://aclanthology.org/W15-0122} {Automatic noun compound
  interpretation using deep neural networks and word embeddings}.
\newblock In \emph{Proceedings of the 11th International Conference on
  Computational Semantics}, pages 173--183, London, UK. Association for
  Computational Linguistics.

\bibitem[{Dozat and Manning(2017)}]{dozat2017deep}
Timothy Dozat and Christopher~D. Manning. 2017.
\newblock \href {http://arxiv.org/abs/1611.01734} {Deep biaffine attention for
  neural dependency parsing}.

\bibitem[{Fares et~al.(2018)Fares, Oepen, and
  Velldal}]{fares-etal-2018-transfer}
Murhaf Fares, Stephan Oepen, and Erik Velldal. 2018.
\newblock \href {https://doi.org/10.18653/v1/D18-1178} {Transfer and multi-task
  learning for noun{--}noun compound interpretation}.
\newblock In \emph{Proceedings of the 2018 Conference on Empirical Methods in
  Natural Language Processing}, pages 1488--1498, Brussels, Belgium.
  Association for Computational Linguistics.

\bibitem[{Fu et~al.(2020)Fu, Tan, Chen, Huang, and Huang}]{fu2020nested}
Yao Fu, Chuanqi Tan, Mosha Chen, Songfang Huang, and Fei Huang. 2020.
\newblock \href {http://arxiv.org/abs/2012.08478} {Nested named entity
  recognition with partially-observed treecrfs}.

\bibitem[{Gharbieh et~al.(2017)Gharbieh, Bhavsar, and
  Cook}]{gharbieh-etal-2017-deep}
Waseem Gharbieh, Virendrakumar Bhavsar, and Paul Cook. 2017.
\newblock \href {https://doi.org/10.18653/v1/S17-1006} {Deep learning models
  for multiword expression identification}.
\newblock In \emph{Proceedings of the 6th Joint Conference on Lexical and
  Computational Semantics (*{SEM} 2017)}, pages 54--64, Vancouver, Canada.
  Association for Computational Linguistics.

\bibitem[{Gooding et~al.(2020)Gooding, Taslimipoor, and
  Kochmar}]{gooding-etal-2020-incorporating}
Sian Gooding, Shiva Taslimipoor, and Ekaterina Kochmar. 2020.
\newblock \href {https://aclanthology.org/2020.readi-1.3} {Incorporating
  multiword expressions in phrase complexity estimation}.
\newblock In \emph{Proceedings of the 1st Workshop on Tools and Resources to
  Empower People with REAding DIfficulties (READI)}, pages 14--19, Marseille,
  France. European Language Resources Association.

\bibitem[{Goyal and Huet(2016)}]{goyal2016}
Pawan Goyal and Gérard Huet. 2016.
\newblock \href {https://doi.org/10.15398/jlm.v4i2.108} {Design and analysis of
  a lean interface for sanskrit corpus annotation}.
\newblock \emph{Journal of Language Modelling}, 4:145.

\bibitem[{Goyal and Kulkarni(2014)}]{goyal-kulkarni-2014-converting}
Pawan Goyal and Amba Kulkarni. 2014.
\newblock \href {https://aclanthology.org/C14-1173} {Converting phrase
  structures to dependency structures in {S}anskrit}.
\newblock In \emph{Proceedings of {COLING} 2014, the 25th International
  Conference on Computational Linguistics: Technical Papers}, pages 1834--1843,
  Dublin, Ireland. Dublin City University and Association for Computational
  Linguistics.

\bibitem[{He et~al.(2018)He, Lee, Levy, and Zettlemoyer}]{he-etal-2018-jointly}
Luheng He, Kenton Lee, Omer Levy, and Luke Zettlemoyer. 2018.
\newblock \href {https://doi.org/10.18653/v1/P18-2058} {Jointly predicting
  predicates and arguments in neural semantic role labeling}.
\newblock In \emph{Proceedings of the 56th Annual Meeting of the Association
  for Computational Linguistics (Volume 2: Short Papers)}, pages 364--369,
  Melbourne, Australia. Association for Computational Linguistics.

\bibitem[{Hellwig and Nehrdich(2018)}]{hellwig-nehrdich-2018-sanskrit}
Oliver Hellwig and Sebastian Nehrdich. 2018.
\newblock \href {https://doi.org/10.18653/v1/D18-1295} {{S}anskrit word
  segmentation using character-level recurrent and convolutional neural
  networks}.
\newblock In \emph{Proceedings of the 2018 Conference on Empirical Methods in
  Natural Language Processing}, pages 2754--2763, Brussels, Belgium.
  Association for Computational Linguistics.

\bibitem[{Hu et~al.(2022)Hu, Yang, Jin, Chen, and
  Xiao}]{hu-etal-2022-improving}
Chengwei Hu, Deqing Yang, Haoliang Jin, Zhen Chen, and Yanghua Xiao. 2022.
\newblock \href {https://aclanthology.org/2022.coling-1.163} {Improving
  continual relation extraction through prototypical contrastive learning}.
\newblock In \emph{Proceedings of the 29th International Conference on
  Computational Linguistics}, pages 1885--1895, Gyeongju, Republic of Korea.
  International Committee on Computational Linguistics.

\bibitem[{Huet(2009)}]{Huet2009SanskritS}
Gérard Huet. 2009.
\newblock \href {https://gallium.inria.fr/~huet/PUBLIC/SALA.pdf} {Sanskrit
  segmentation}.
\newblock In \emph{Proceedings of the South Asian Languages Analysis Roundtable
  XXVIII}, Denton, Ohio.

\bibitem[{Krishna et~al.(2016)Krishna, Satuluri, Sharma, Kumar, and
  Goyal}]{krishna-etal-2016-compound}
Amrith Krishna, Pavankumar Satuluri, Shubham Sharma, Apurv Kumar, and Pawan
  Goyal. 2016.
\newblock \href {https://aclanthology.org/W16-3701} {Compound type
  identification in {S}anskrit: What roles do the corpus and grammar play?}
\newblock In \emph{Proceedings of the 6th Workshop on South and Southeast
  {A}sian Natural Language Processing ({WSSANLP}2016)}, pages 1--10, Osaka,
  Japan. The COLING 2016 Organizing Committee.

\bibitem[{Kulkarni(2019)}]{inbook-amba}
Amba Kulkarni. 2019.
\newblock \href {https://dkprintworld.com/product/sanskrit-parsing/}
  {\emph{Sanskrit Parsing based on the Theories of Shabdabodha}}. D. K.
  PrintWorld and Indian Institute of Advanced Study.

\bibitem[{Kulkarni(2021)}]{kulkarni-acm}
Amba Kulkarni. 2021.
\newblock \href {https://doi.org/https://doi.org/10.1145/3418061} {Sanskrit
  parsing following indian theories of verbal cognition}.
\newblock In \emph{ACM Transactions on Asian and Low-Resource Language
  Information Processing}, pages 1--38.

\bibitem[{Kulkarni and Kumar(2011)}]{Kulkarni2011StatisticalCP}
Amba Kulkarni and Anil Kumar. 2011.
\newblock \href
  {https://sanskrit.uohyd.ac.in/faculty/amba/PUBLICATIONS/papers/samaasa_const_parser_icon2011.pdf}
  {Statistical constituency parser for sanskrit compounds}.
\newblock In \emph{Proceedings of ICON-2011: 9th International Conference on
  Natural Language Processing}, International Institute of Information
  Technology, Hyderabad, India. Macmillan Publishers.

\bibitem[{Kulkarni and Kumar(2013)}]{kulkarni2013}
Amba Kulkarni and Anil Kumar. 2013.
\newblock \href
  {https://dkprintworld.com/product/recent-researches-in-sanskrit-computational-linguistics/}
  {Clues from a\d{s}t\={a}dhy\={a}y\={\i} for compound type identification}.
\newblock In \emph{Proceedings of the 5th International Sanskrit Computational
  Linguistics Symposium}, IIT Bombay, India. D. K. PrintWorld and Indian
  Institute of Advanced Study.

\bibitem[{Kulkarni and Sharma(2019)}]{kulkarni-sharma-2019-paninian}
Amba Kulkarni and Dipti Sharma. 2019.
\newblock \href {https://doi.org/10.18653/v1/W19-7724} {{P}{\=a}ṇinian
  syntactico-semantic relation labels}.
\newblock In \emph{Proceedings of the Fifth International Conference on
  Dependency Linguistics (Depling, SyntaxFest 2019)}, pages 198--208, Paris,
  France. Association for Computational Linguistics.

\bibitem[{Kumar(2012)}]{anil_thesis}
Anil Kumar. 2012.
\newblock \href
  {https://sanskrit.uohyd.ac.in/faculty/amba/PUBLICATIONS/Student_Thesis/anil.pdf}
  {\emph{An automatic Sanskrit compound processing}}.
\newblock Ph.D. thesis, University of Hyderabad.

\bibitem[{Kumar et~al.(2009)Kumar, Sheeba, and Kulkarni}]{Kumar2009SanskritCP}
Anil Kumar, V~Sheeba, and Amba Kulkarni. 2009.
\newblock \href
  {https://sanskrit.uohyd.ac.in/faculty/amba/PUBLICATIONS/papers/paper4icon.pdf}
  {Sanskrit compound paraphrase generator}.
\newblock In \emph{Proceedings of the ICON 2009: 7th International Conference
  on Natural Language Processing}, University of Hyderabad, Hyderabad.
  Macmillan Publishers.

\bibitem[{Lapata and Keller(2004)}]{lapata-keller-2004-web}
Mirella Lapata and Frank Keller. 2004.
\newblock \href {https://aclanthology.org/N04-1016} {The web as a baseline:
  Evaluating the performance of unsupervised web-based models for a range of
  {NLP} tasks}.
\newblock In \emph{Proceedings of the Human Language Technology Conference of
  the North {A}merican Chapter of the Association for Computational
  Linguistics: {HLT}-{NAACL} 2004}, pages 121--128, Boston, Massachusetts, USA.
  Association for Computational Linguistics.

\bibitem[{Lou et~al.(2022)Lou, Yang, and Tu}]{lou-etal-2022-nested}
Chao Lou, Songlin Yang, and Kewei Tu. 2022.
\newblock \href {https://doi.org/10.18653/v1/2022.acl-long.428} {Nested named
  entity recognition as latent lexicalized constituency parsing}.
\newblock In \emph{Proceedings of the 60th Annual Meeting of the Association
  for Computational Linguistics (Volume 1: Long Papers)}, pages 6183--6198,
  Dublin, Ireland. Association for Computational Linguistics.

\bibitem[{Ma et~al.(2018)Ma, Hu, Liu, Peng, Neubig, and
  Hovy}]{ma-etal-2018-stack}
Xuezhe Ma, Zecong Hu, Jingzhou Liu, Nanyun Peng, Graham Neubig, and Eduard
  Hovy. 2018.
\newblock \href {https://doi.org/10.18653/v1/P18-1130} {Stack-pointer networks
  for dependency parsing}.
\newblock In \emph{Proceedings of the 56th Annual Meeting of the Association
  for Computational Linguistics (Volume 1: Long Papers)}, pages 1403--1414,
  Melbourne, Australia. Association for Computational Linguistics.

\bibitem[{Maurya and Bahadur(2022)}]{inbook-archana}
Archana Maurya and Promila Bahadur. 2022.
\newblock \href {https://doi.org/10.1007/978-981-16-3346-1_56} {\emph{A
  Detailed Analysis of Word Sense Disambiguation Algorithms and Approaches for
  Indian Languages}}, pages 693--710.

\bibitem[{Maurya et~al.(2023)Maurya, Bahadur, and
  Garg}]{10.1007/978-981-19-3148-2_40}
Archana~Sachindeo Maurya, Promila Bahadur, and Srishti Garg. 2023.
\newblock Approach toward word sense disambiguation for the english-to-sanskrit
  language using na{\"i}ve bayesian classification.
\newblock In \emph{Proceedings of Third Doctoral Symposium on Computational
  Intelligence}, pages 477--491, Singapore. Springer Nature Singapore.

\bibitem[{Mittal(2010)}]{mittal-2010-automatic}
Vipul Mittal. 2010.
\newblock \href {https://aclanthology.org/P10-3015} {Automatic {S}anskrit
  segmentizer using finite state transducers}.
\newblock In \emph{Proceedings of the {ACL} 2010 Student Research Workshop},
  pages 85--90, Uppsala, Sweden. Association for Computational Linguistics.

\bibitem[{Nadgeri et~al.(2021)Nadgeri, Bastos, Singh, Mulang{'}, Hoffart,
  Shekarpour, and Saraswat}]{nadgeri-etal-2021-kgpool}
Abhishek Nadgeri, Anson Bastos, Kuldeep Singh, Isaiah~Onando Mulang{'},
  Johannes Hoffart, Saeedeh Shekarpour, and Vijay Saraswat. 2021.
\newblock \href {https://doi.org/10.18653/v1/2021.findings-acl.48} {{KGP}ool:
  Dynamic knowledge graph context selection for relation extraction}.
\newblock In \emph{Findings of the Association for Computational Linguistics:
  ACL-IJCNLP 2021}, pages 535--548, Online. Association for Computational
  Linguistics.

\bibitem[{Nguyen et~al.(2021)Nguyen, Lai, Pouran Ben~Veyseh, and
  Nguyen}]{nguyen-etal-2021-trankit}
Minh~Van Nguyen, Viet~Dac Lai, Amir Pouran Ben~Veyseh, and Thien~Huu Nguyen.
  2021.
\newblock \href {https://doi.org/10.18653/v1/2021.eacl-demos.10} {Trankit: A
  light-weight transformer-based toolkit for multilingual natural language
  processing}.
\newblock In \emph{Proceedings of the 16th Conference of the European Chapter
  of the Association for Computational Linguistics: System Demonstrations},
  pages 80--90, Online. Association for Computational Linguistics.

\bibitem[{Pa\d{n}ini(500 BCE)}]{panini}
Pa\d{n}ini. 500 BCE.
\newblock \href {http://panini.phil.hhu.de/panini/panini/}
  {A\d{s}t\={a}dhy\={a}y\={i}}.

\bibitem[{Ponkiya et~al.(2021)Ponkiya, Kanojia, Bhattacharyya, and
  Palshikar}]{ponkiya-etal-2021-framenet}
Girishkumar Ponkiya, Diptesh Kanojia, Pushpak Bhattacharyya, and Girish
  Palshikar. 2021.
\newblock \href {https://doi.org/10.18653/v1/2021.findings-acl.256}
  {{F}rame{N}et-assisted noun compound interpretation}.
\newblock In \emph{Findings of the Association for Computational Linguistics:
  ACL-IJCNLP 2021}, pages 2901--2911, Online. Association for Computational
  Linguistics.

\bibitem[{Ponkiya et~al.(2020)Ponkiya, Murthy, Bhattacharyya, and
  Palshikar}]{ponkiya-etal-2020-looking}
Girishkumar Ponkiya, Rudra Murthy, Pushpak Bhattacharyya, and Girish Palshikar.
  2020.
\newblock \href {https://doi.org/10.18653/v1/2020.findings-emnlp.386} {Looking
  inside noun compounds: Unsupervised prepositional and free paraphrasing}.
\newblock In \emph{Findings of the Association for Computational Linguistics:
  EMNLP 2020}, pages 4313--4323, Online. Association for Computational
  Linguistics.

\bibitem[{Ponkiya et~al.(2018)Ponkiya, Patel, Bhattacharyya, and
  Palshikar}]{ponkiya-etal-2018-treat}
Girishkumar Ponkiya, Kevin Patel, Pushpak Bhattacharyya, and Girish Palshikar.
  2018.
\newblock \href {https://aclanthology.org/C18-1155} {Treat us like the
  sequences we are: Prepositional paraphrasing of noun compounds using {LSTM}}.
\newblock In \emph{Proceedings of the 27th International Conference on
  Computational Linguistics}, pages 1827--1836, Santa Fe, New Mexico, USA.
  Association for Computational Linguistics.

\bibitem[{Premasiri and Ranasinghe(2022)}]{premasiri2022berts}
Damith Premasiri and Tharindu Ranasinghe. 2022.
\newblock \href {http://arxiv.org/abs/2208.07832} {Bert(s) to detect multiword
  expressions}.

\bibitem[{Sandhan et~al.(2022{\natexlab{a}})Sandhan, Gupta, Terdalkar, Sandhan,
  Samanta, Behera, and Goyal}]{sandhan-etal-2022-novel}
Jivnesh Sandhan, Ashish Gupta, Hrishikesh Terdalkar, Tushar Sandhan, Suvendu
  Samanta, Laxmidhar Behera, and Pawan Goyal. 2022{\natexlab{a}}.
\newblock \href {https://aclanthology.org/2022.coling-1.358} {A novel
  multi-task learning approach for context-sensitive compound type
  identification in {S}anskrit}.
\newblock In \emph{Proceedings of the 29th International Conference on
  Computational Linguistics}, pages 4071--4083, Gyeongju, Republic of Korea.
  International Committee on Computational Linguistics.

\bibitem[{Sandhan et~al.(2019)Sandhan, Krishna, Goyal, and
  Behera}]{sandhan-etal-2019-revisiting}
Jivnesh Sandhan, Amrith Krishna, Pawan Goyal, and Laxmidhar Behera. 2019.
\newblock \href {https://aclanthology.org/W19-7503} {Revisiting the role of
  feature engineering for compound type identification in {S}anskrit}.
\newblock In \emph{Proceedings of the 6th International Sanskrit Computational
  Linguistics Symposium}, pages 28--44, IIT Kharagpur, India. Association for
  Computational Linguistics.

\bibitem[{Sandhan et~al.(2022{\natexlab{b}})Sandhan, Singha, Rao, Samanta,
  Behera, and Goyal}]{sandhan-etal-2022-translist}
Jivnesh Sandhan, Rathin Singha, Narein Rao, Suvendu Samanta, Laxmidhar Behera,
  and Pawan Goyal. 2022{\natexlab{b}}.
\newblock \href {https://aclanthology.org/2022.findings-emnlp.513}
  {{T}rans{LIST}: A transformer-based linguistically informed {S}anskrit
  tokenizer}.
\newblock In \emph{Findings of the Association for Computational Linguistics:
  EMNLP 2022}, pages 6902--6912, Abu Dhabi, United Arab Emirates. Association
  for Computational Linguistics.

\bibitem[{Satuluri and Kulkarni(2013)}]{pavan2013}
Pavankumar Satuluri and Amba Kulkarni. 2013.
\newblock \href {https://ltrc.iiit.ac.in/icon/2013/proceedings.php} {Generation
  of sanskrit compounds}.
\newblock In \emph{Proceedings of ICON-2013: 10th International Conference on
  Natural Language Processing}, CDAC Noida, India.

\bibitem[{Shwartz and Waterson(2018)}]{shwartz-waterson-2018-olive}
Vered Shwartz and Chris Waterson. 2018.
\newblock \href {https://doi.org/10.18653/v1/N18-2035} {Olive oil is made
  \textit{of} olives, baby oil is made \textit{for} babies: Interpreting noun
  compounds using paraphrases in a neural model}.
\newblock In \emph{Proceedings of the 2018 Conference of the North {A}merican
  Chapter of the Association for Computational Linguistics: Human Language
  Technologies, Volume 2 (Short Papers)}, pages 218--224, New Orleans,
  Louisiana. Association for Computational Linguistics.

\bibitem[{Terdalkar and
  Bhattacharya(2019)}]{terdalkar-bhattacharya-2019-framework}
Hrishikesh Terdalkar and Arnab Bhattacharya. 2019.
\newblock \href {https://aclanthology.org/W19-7508} {Framework for
  question-answering in {S}anskrit through automated construction of knowledge
  graphs}.
\newblock In \emph{Proceedings of the 6th International Sanskrit Computational
  Linguistics Symposium}, pages 97--116, IIT Kharagpur, India. Association for
  Computational Linguistics.

\bibitem[{Tian et~al.(2021)Tian, Chen, Song, and
  Wan}]{tian-etal-2021-dependency}
Yuanhe Tian, Guimin Chen, Yan Song, and Xiang Wan. 2021.
\newblock \href {https://doi.org/10.18653/v1/2021.acl-long.344}
  {Dependency-driven relation extraction with attentive graph convolutional
  networks}.
\newblock In \emph{Proceedings of the 59th Annual Meeting of the Association
  for Computational Linguistics and the 11th International Joint Conference on
  Natural Language Processing (Volume 1: Long Papers)}, pages 4458--4471,
  Online. Association for Computational Linguistics.

\bibitem[{Wang et~al.(2022)Wang, Song, Liu, Gao, Lin, Cao, and
  Sui}]{wang2022more}
Peiyi Wang, Yifan Song, Tianyu Liu, Rundong Gao, Binghuai Lin, Yunbo Cao, and
  Zhifang Sui. 2022.
\newblock \href {http://arxiv.org/abs/2209.00243} {Less is more: Rethinking
  state-of-the-art continual relation extraction models with a frustratingly
  easy but effective approach}.

\bibitem[{Yan et~al.(2021)Yan, Gui, Dai, Guo, Zhang, and Qiu}]{yan2021unified}
Hang Yan, Tao Gui, Junqi Dai, Qipeng Guo, Zheng Zhang, and Xipeng Qiu. 2021.
\newblock \href {http://arxiv.org/abs/2106.01223} {A unified generative
  framework for various ner subtasks}.

\bibitem[{Yang and Tu(2021)}]{yang2021bottomup}
Songlin Yang and Kewei Tu. 2021.
\newblock \href {http://arxiv.org/abs/2110.05419} {Bottom-up constituency
  parsing and nested named entity recognition with pointer networks}.

\bibitem[{Yuan et~al.(2022)Yuan, Tan, Huang, and Huang}]{yuan2022fusing}
Zheng Yuan, Chuanqi Tan, Songfang Huang, and Fei Huang. 2022.
\newblock \href {http://arxiv.org/abs/2110.07480} {Fusing heterogeneous factors
  with triaffine mechanism for nested named entity recognition}.

\bibitem[{Zhang et~al.(2022)Zhang, Jindal, and
  Li}]{zhang-etal-2022-label-definitions}
Li~Zhang, Ishan Jindal, and Yunyao Li. 2022.
\newblock \href {https://doi.org/10.18653/v1/2022.naacl-main.411} {Label
  definitions improve semantic role labeling}.
\newblock In \emph{Proceedings of the 2022 Conference of the North American
  Chapter of the Association for Computational Linguistics: Human Language
  Technologies}, pages 5613--5620, Seattle, United States. Association for
  Computational Linguistics.

\bibitem[{Ziering and van~der Plas(2015)}]{ziering-van-der-plas-2015-distance}
Patrick Ziering and Lonneke van~der Plas. 2015.
\newblock \href {https://aclanthology.org/W15-0112} {From a distance: Using
  cross-lingual word alignments for noun compound bracketing}.
\newblock In \emph{Proceedings of the 11th International Conference on
  Computational Semantics}, pages 82--87, London, UK. Association for
  Computational Linguistics.

\end{thebibliography}
\bibliographystyle{acl_natbib}

\newpage
\appendix

\section{Why NeCTI as a Dependency Parsing Task?}
\label{Why_dp}
Compounds in language are semantic constructions. While a limited number of rules derived from Sanskrit Grammar aid in determining the syntactic structure of a compound, they offer limited assistance in uncovering its meaning. The meaning of a compound primarily arises from the semantic relationship between its components. This semantic relationship, known as ``sāmarthya,'' is expressed through various types of semantic compounds. These compound types also facilitate the identification of the headword within a compound. The headword can be one of the constituents or an entirely distinct word modified by the resulting compound. Compounds consisting of more than two components are typically formed through successive binary combinations of the components, following specific relations. As a result, a nested structure of binary compounds is created, with a few exceptions. Therefore, the semantic compound types play a vital role in determining the correct nested structure amidst the various possible structures of a compound. Each of these structures represents a constituency span, and a parser equipped with compound type identification assists in accurately identifying the appropriate constituency span.

\begin{figure}[!h]
\begin{center}
\begin{subfigure}{.5\linewidth}
 \centering
 \includegraphics[width=0.5\textwidth]{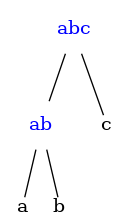}
 \caption{< < a - b > - c >}
 \label{fig:1sub1}
\end{subfigure}%
\begin{subfigure}{.5\linewidth}
 \centering
 \includegraphics[width=0.5\textwidth]{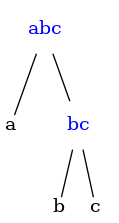}
 \caption{< a - < b - c > >}
 \label{fig:1sub2}
\end{subfigure}
\caption{Possible constituency spans for a three component compound a-b-c}
\label{fig:cspan1}
\end{center}
\end{figure}

Treating this as a constituency parsing task presents several challenges. Figure \ref{fig:cspan1} illustrates the potential constituency spans  for a 3-component compound $a-b-c$, with $ab$ and $bc$ representing the intermediate compounds. First, the nested structure of compounds does not conform to a syntactic structure; instead, it follows a semantic structure based on the relations between the components rather than their position or relative co-occurrence. Second, the intermediate compounds within the structure are not categories but stem forms. Each intermediate compound serves as an entity that modifies the meanings of one of its constituents, subsequently combining with another component to form a larger compound.

\begin{figure}[!h]
\begin{center}
\begin{subfigure}{.5\linewidth}
 \centering
 \includegraphics[width=0.5\textwidth]{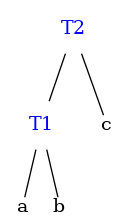}
 \caption{< < a - b > T1 - c > T2}
 \label{fig:2sub1}
\end{subfigure}%
\begin{subfigure}{.5\linewidth}
 \centering
 \includegraphics[width=0.5\textwidth]{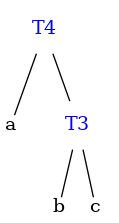}
 \caption{< a - < b - c > T3 > T4}
 \label{fig:2sub2}
\end{subfigure}
\caption{Possible constituency spans for a three component compound a-b-c with semantic types}
\label{fig:cspan2}
\end{center}
\end{figure}

\newcite{anil_thesis} formulated this as constituency parsing, where the identification of semantic compatibility was performed based on the relative co-occurrence and relative position of the components. As shown in Figure \ref{fig:cspan2}, the intermediate compounds, which lacked additional information, were substituted with their respective semantic compound types in the type identification stage. 

The semantic types serve as semantic constructions for compounds, rather than being syntactic categories and play a crucial role in determining the meaning of the compound. For example the compound \textit{vidy\={a}layagha\d{n}\d{t}\={a}} has the nested structure <<\textit{vidy\={a}-\={a}laya}>$T6$-\textit{gha\d{n}\d{t}\={a}}>$T6$. The type T6 indicates \textit{\d{s}a\d{s}\d{t}h\={i} tatpuru\d{s}a} compound ($6^{th}$ case determinative compound) inferring a possessive relationship. \textit{Vidy\={a}} (knowledge) and \textit{\={a}laya} (place) combine to form the intermediate compound \textit{vidy\={a}laya} (the place of knowledge, viz. school), which combines with \textit{gha\d{n}\d{t}\={a}} (bell) to form the whole compound indicating school-bell. The possessive relation expressed by the compound type (T6) is akin to a dependency relation, and this holds true for other compound types as well. 

\begin{figure}[!h]
\begin{center}
\begin{subfigure}{.7\linewidth}
 \centering
 \includegraphics[width=1\textwidth]{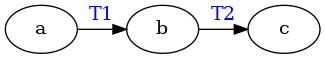}
 \caption{$<<a-b>T1-c>T2$}
 \label{fig:3sub1}
\end{subfigure}
\begin{subfigure}{.7\linewidth}
 \centering
 \includegraphics[width=1\textwidth]{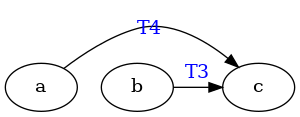}
 \caption{$<a-<b-c>T3>T4$}
 \label{fig:3sub2}
\end{subfigure}
\caption{Dependency Structure for the three component compound a-b-c with semantic types as labels}
\label{fig:dep}
\end{center}
\end{figure}

Consequently, when represented with compound types, the same constituency spans can be depicted as dependency structures by annotating the types as directed relations and removing the extra nodes indicating the types. The dependency structures for the previous example ($a-b-c$) are shown in Figure \ref{fig:dep}. Furthermore, the head words are not explicitly marked in constituency spans and can only be identified through their corresponding types. However, with dependency structures, the head word can be determined by the labels directed towards it within the compound.  Notably, these dependency structures faithfully capture the constituency information and are mutually convertible with their corresponding constituency spans.

There are several considerations behind the decision to treat NeCTI as a dependency parsing task. 
First, the semantic relations among compound components resemble dependency relations and can be represented as directed labels within the dependency parse structure. Second, this approach concisely represents the semantic relations between compound components without introducing extra intermediary nodes. Lastly, it enables the simultaneous identification of the structure or constituency span alongside the identification of compound types.

\section{Experiment Details}
\label{details}
\paragraph{Hyper-parameters:} 
For our proposed system, we build on the top of the codebase from BiAFF, as developed by \newcite{ma-etal-2018-stack}. We configure the hyperparameters as follows: a batch size of 16, 100 training iterations, a dropout rate of 0.33, 2 stacked Bi-LSTM layers, a learning rate of 0.002, and the remaining parameters set identically to those used in the work of \newcite{ma-etal-2018-stack}. We use manual tuning for the hyper-parameter selection and F1-score criteria on dev set's performance .  Our codebase is publicly available and released under a creative-common license. We use FastText word embeddings for the proposed framework.

\paragraph{Computing Infrastructure Used:}
We perform our experiments using a single GPU equipped with an NVIDIA A40, 48 GB GPU memory, and 10752 CUDA cores. We employ a single GPU with an NVIDIA Quadro RTX 4000, 8 GB GPU memory, and 2304 CUDA cores for our proposed system.

\end{document}